\documentclass{article} 
\usepackage[final]{colm2026_conference}

\usepackage{microtype}
\usepackage{hyperref}
\usepackage{url}
\usepackage{booktabs}
\usepackage{graphicx}
\usepackage{booktabs}
\usepackage{amsmath}
\usepackage{amssymb}
\usepackage{float}
\usepackage{algorithm}
\usepackage{algpseudocode}
\usepackage{multirow} 
\usepackage{wrapfig}

\usepackage{pifont}

\usepackage{colortbl}
\usepackage{xcolor}
\usepackage[dvipsnames]{xcolor}
\usepackage{cleveref}
\usepackage{enumitem}
\usepackage{caption}
\usepackage{wrapfig}

\definecolor{eccvblue}{rgb}{0.2,0.5,0.8}
\definecolor{teal}{rgb}{0.0, 0.5, 0.5}

\usepackage{pifont}
\renewcommand{\paragraph}[1]{\vspace{1.25mm}\noindent\textbf{#1}}

\definecolor{baselinecolor}{gray}{.95}

\makeatletter
\def\blfootnote{\xdef\@thefnmark{}\@footnotetext}
\makeatother

\usepackage{lineno}

\definecolor{darkblue}{rgb}{0, 0, 0.5}
\hypersetup{colorlinks=true, citecolor=darkblue, linkcolor=darkblue, urlcolor=darkblue,
  pdftitle={The Cost of Language: Centroid Erasure Exposes and Exploits Modal Competition in Multimodal Language Models},
  pdfauthor={Akshay Paruchuri, Ishan Chatterjee, Henry Fuchs, Ehsan Adeli, Piotr Didyk}}

\title{The Cost of Language: Centroid Erasure Exposes and Exploits Modal Competition in Multimodal Language Models}

\author{Akshay Paruchuri$^{1,4}$, Ishan Chatterjee$^2$, Henry Fuchs$^3$, Ehsan Adeli$^4$, Piotr Didyk$^1$ \\
$^1$Università della Svizzera italiana, $^2$University of Washington \\
$^3$UNC Chapel Hill, $^4$Stanford University \\
\texttt{akshaypa@stanford.edu}
}

\begin{document}

\ifcolmsubmission
\linenumbers
\fi

\maketitle

\vspace{-1em}
\begin{figure*}[h]
    \centering
    \includegraphics[width=0.95\textwidth]{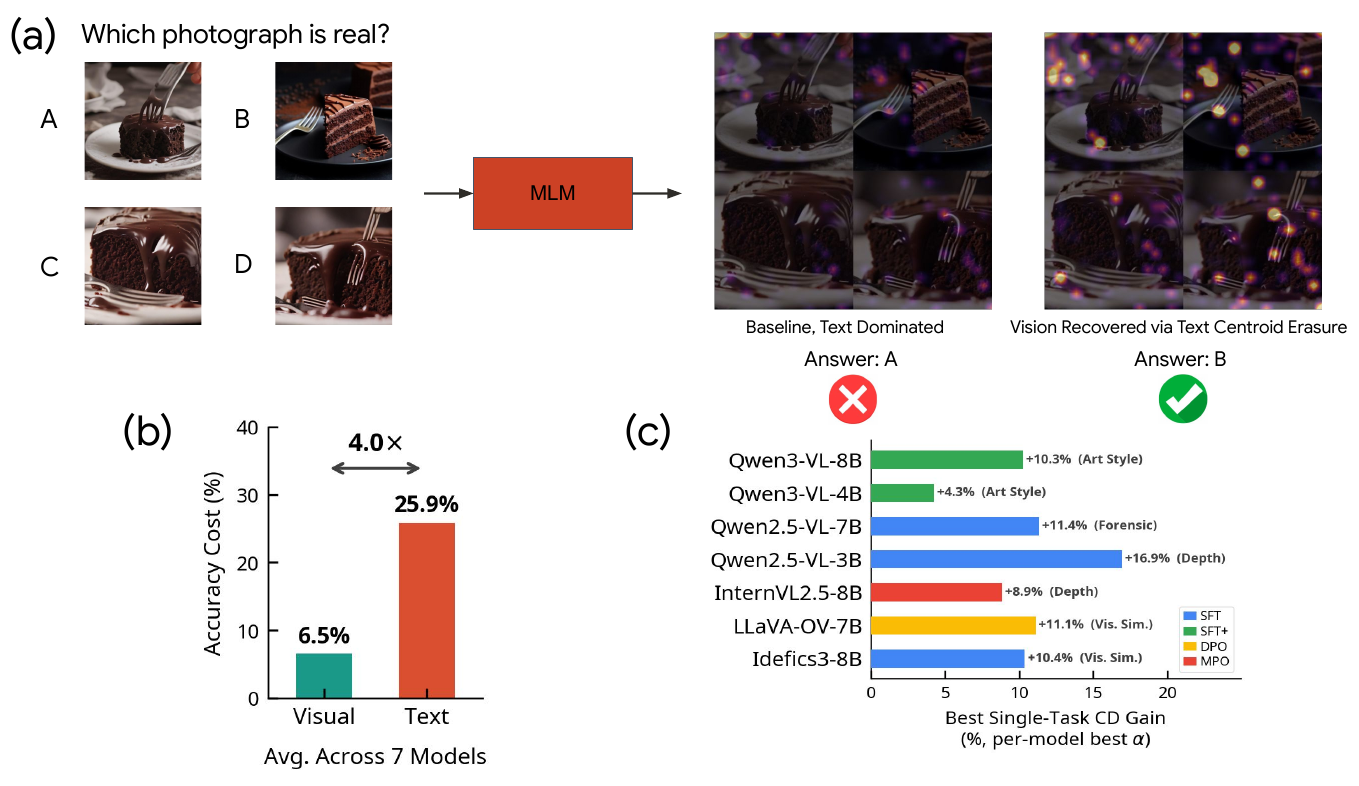}
    \vspace{-1.25em}
    \caption{\textbf{Centroid erasure exposes and exploits modal competition in multimodal language models.} \textbf{(a)}~In one BLINK example, text centroid replacement raises visual attention at L22 (3.6\%\,$\to$\,11.1\%), while text centroid contrastive decoding (TCCD) flips the prediction from incorrect to correct. The displayed heatmaps are qualitative; quantitative attention is computed separately after RoPE. \textbf{(b)}~Across seven models and six BLINK tasks, post-image text replacement costs 4$\times$ more accuracy than visual replacement (25.9\% vs.\ 6.5\%). Since text replacement also changes the task and answer interface, the gap measures relative dependence. \textbf{(c)}~At oracle best-per-task $\alpha_\text{interp}$, TCCD recovers up to +16.9\% accuracy on an individual task, with every model showing positive gains on at least one task.}
    \label{fig:overview}
\end{figure*}

\begin{abstract}
Multimodal language models systematically underperform on visual perception tasks, yet the structure underlying this failure remains poorly understood. We propose centroid replacement, mapping tokens to their nearest K-means centroid and removing within-cluster residual structure, as a controlled probe for modal dependence. Across seven models spanning four architecture families, post-image text replacement costs 4$\times$ more accuracy than visual replacement on six BLINK tasks; VPBench and MedBLINK preserve this aggregate ordering. Since text replacement also disrupts the task and answer interface, the gap measures relative dependence. Text centroid contrastive decoding (TCCD) exploits this asymmetry without retraining, recovering up to +16.9\% accuracy on an individual task at oracle best-per-task $\alpha_\text{interp}$; every model gains on at least one task, although gains vary across tasks and models. Together, centroid replacement provides a retraining-free audit of modal dependence on labeled data, while TCCD offers a reusable inference-time intervention. Code and artifacts are available at: \url{https://github.com/yahskapar/centroid-erasure}
\end{abstract}

\section{Introduction}
\label{sec:intro}

Multimodal language models (MLMs) extend the capabilities of large language models (LLMs) to visual inputs, yet they systematically underperform on tasks that require visual perception. On benchmarks such as BLINK~\citep{fu2024blink} (depth estimation, visual similarity, forensic detection), even state-of-the-art MLMs fall short of human perceptual capabilities and of what their own vision encoders can achieve in isolation~\citep{tong2024eyes, fu2025hiddenplainsightvlms}. Recent work has traced this failure to the language model itself: visual representations encode the information needed to solve these tasks, but the LLM backbone fails to use it~\citep{fu2025hiddenplainsightvlms}. Together, these findings point to a modal imbalance favoring text~\citep{deng2025words, wu2025mitigating}, further reflected in gradient and attention skew toward text.

These perceptual failures carry substantial consequences as MLMs are deployed in high-stakes domains. In healthcare~\citep{tu2024towards}, frontier MLMs can generate detailed clinical findings for images that were never provided while achieving top chest X-ray question-answering scores without visual input~\citep{asadi2026mirage}. Such ungrounded perception errors can propagate into triage, diagnosis, and treatment planning. The problem is further compounded in agentic systems: personal-health architectures rely on each perception step being faithful to the input~\citep{heydari2025anatomy}, and a single instance of text-dominated perception can propagate through downstream reasoning in single-agent systems (e.g.,~\citealp{merrill2026transforming}) and multi-agent pipelines~\citep{kim2024mdagents}. Measuring how predictions depend on language versus visual structure, and testing whether targeted inference-time interventions help without retraining, is therefore not merely an academic question but a practical requirement for trustworthy multimodal deployment.

Recently, NIST AI 800-4~\citep{rao2026challenges} identified functionality monitoring as an open challenge for deployed AI systems, ensuring that a model's outputs are grounded in the intended evidence channels rather than merely appearing correct on benchmark metrics. Existing approaches generally require labeled data, human review, or task-specific telemetry. A retraining-free forward-pass probe that quantifies relative dependence on fine-grained text versus visual structure on a labeled audit set can complement these approaches.

In this work, we propose \emph{centroid replacement} as a controlled probe for modal dependence: collapsing each token to its nearest K-means centroid and measuring the resulting accuracy cost. This removes within-cluster residual structure while retaining the centroids themselves; throughout, we call this residual removal ``centroid erasure''. Applying the same replacement rule separately to visual and post-image text tokens reveals a stark asymmetry. Across seven models spanning four architecture families, text replacement costs 4$\times$ more accuracy than visual replacement (25.9\% vs.\ 6.5\% on average; \Cref{fig:overview}b). We call the resulting dependence asymmetry \emph{modal competition}. These tasks place answer-discriminating evidence in the image while text supplies the task, candidate semantics, and answer interface; because text replacement also disrupts that interface, the probe measures relative dependence rather than a causal effect of harmful language competition.

We further exploit this asymmetry through \emph{text centroid contrastive decoding} (TCCD): running a second forward pass with post-image text residual structure removed and contrastively decoding the original output against this reference. At oracle best-per-task $\alpha_\text{interp}$, TCCD recovers up to +16.9\% accuracy on an individual task, and every model gains on at least one task, although gains vary across tasks and models (\Cref{fig:overview}c). We test specificity with an equal-norm matched-random control and a directional sweep of contrastive strength $\alpha_\text{cd}$ (\Cref{sec:mechanism}). Beyond BLINK, separately scored VPBench and MedBLINK portfolios retain the aggregate text--visual ordering, while recovery is more variable (\Cref{sec:cross_benchmark}). An exploratory checkpoint comparison gives a larger oracle best-per-task mean gain for SFT/DPO-labeled models (+5.6\%) than for SFT+/MPO models (+1.5\%); fixed and cross-validated protocols preserve this ordering, but architecture, data, baseline accuracy, and task composition remain confounded.

\clearpage

Our contributions are:
\begin{enumerate}[leftmargin=*, itemsep=2pt, topsep=4pt]
    \item \textbf{Revealing a consistent asymmetry in modal dependence.} Across seven models and four architecture families, removing post-image text residual structure costs 4$\times$ more accuracy than removing visual residual structure; separately scored VPBench and MedBLINK portfolios preserve the aggregate ordering (\Cref{sec:measuring}).
    \item \textbf{Providing a retraining-free inference-time intervention.} In the seven-model, six-task BLINK core, TCCD recovers up to +16.9\% accuracy on an individual task at oracle best-per-task $\alpha_\text{interp}$. The negative-$\alpha_\text{cd}$ side of a directional sweep is monotonic on five of six tasks, and an equal-norm matched-random control tests specificity (\Cref{sec:correcting}).
    \item \textbf{Auditing modal dependence on labeled data.} Positional-span and layer ablations characterize dose- and depth-dependent sensitivity, while an exploratory stage probe tracks visual replacement cost from 22\% at the encoder to 2\% by L16 (\Cref{sec:mechanism}). Together, replacement cost and TCCD response provide retraining-free audit measures, with TCCD response treated as secondary (\Cref{sec:diagnostic}).
\end{enumerate}

Code, centroid artifacts, and aggregate results are available at: \url{https://github.com/yahskapar/centroid-erasure}

\section{Related Work}
\label{sec:related}

\paragraph{Modal imbalance in multimodal language models.} The observation that MLMs under-utilize visual information has emerged from multiple independent lines of evidence. \citet{fu2025hiddenplainsightvlms} showed that vision encoders encode the information needed for visual perception tasks, but the LLM backbone fails to use it; \citet{deng2025words} confirmed text dominance across VLM families, and \citet{shahgir2026vlms} showed that VLMs favor semantic anchors over visual detail. \citet{li2025mbq} found that language token features carry over 10$\times$ the average absolute gradient of vision tokens, while \citet{wu2025mitigating} identified cross-modal attention imbalance and proposed a training-time fix. Earlier work traced modality gaps to contrastive pretraining dynamics~\citep{liang2022mind} and under-optimized unimodal representations~\citep{peng2022balanced, laurenccon2024matters, jiang2025rethinking}. Our work complements these findings with a geometric measure of text--visual dependence and a task- and model-dependent training-free intervention.

\paragraph{Contrastive decoding for multimodal language models.} Contrastive decoding methods typically mitigate hallucinations by subtracting a degraded reference distribution from the model's output logits. VCD~\citep{leng2024mitigating} contrasts original and noise-distorted visual inputs, while LCD~\citep{manevich2024mitigating} subtracts the text-only LLM distribution. DoLa~\citep{chuang2023dola} contrasts later- and earlier-layer logits, SDCD~\citep{xia2026sdcd} shuffles visual patches, and OPERA~\citep{huang2024opera} penalizes over-trusted attention patterns during beam search. However, \citet{yin2025mirage} showed that gains on POPE~\citep{li2023evaluating} are largely distributional artifacts rather than genuine hallucination suppression, questioning CD's role as a universal intervention. TCCD differs in its intervention locus: it constructs the degraded reference by replacing text-side representations at an intermediate layer, then contrasts the resulting output logits. We treat variation in its gains as a secondary post-hoc measure of how strongly the measured dependence responds to this intervention, not as a universal correction or a training objective (\Cref{tab:method_positioning}, \Cref{app:positioning}).

\paragraph{Representation geometry in language models.} Recent work has revealed rich geometric structure in LLM representations. \citet{huh2024platonic} argued that vision and language models converge toward a shared representation, \citet{lee2025shared} found shared global and local geometry across LLM token embeddings, and \citet{park2024geometry} showed that categorical concepts form polytopes in representation space. In the multimodal setting, \citet{papadimitriou2025interpreting} found a small stable dictionary of unimodal concepts that form cross-modal bridges in CLIP and SigLIP embedding spaces. \citet{qi2025beyond} found that randomly permuting vision token embeddings causes only 0.2--2.7\% performance drops, suggesting that VLMs treat visual tokens as a semantic bag of tokens. Token compression methods such as FastV~\citep{chen2024image} and VisionZip~\citep{yang2025visionzip} similarly exploit visual token redundancy. Our centroid analysis connects to this work by applying K-means decomposition to \emph{both} visual and text tokens within the same model, directly measuring a cross-modal asymmetry in dependence on within-cluster structure.

\section{Measuring Modal Competition}
\label{sec:measuring}

We propose \emph{centroid replacement} as a controlled probe for modal dependence. For each model, we fit $K$-means centroids ($K$=256; sensitivity to $K$ and $N$ in \Cref{app:ksweep}) on hidden-state activations from one intermediate layer per modality (text L12, visual L16; \Cref{app:centroids}). The fitting pool contains $N$=2{,}000 MS-COCO images and uses no BLINK labels or evaluation prompts. Centroid replacement interpolates each token toward its nearest centroid: $\mathbf{x}' = \boldsymbol{\mu}_k + \alpha_\text{interp} \cdot (\mathbf{x} - \boldsymbol{\mu}_k)$. At $\alpha_\text{interp}$=0, every token collapses to its cluster center and all within-cluster variation is discarded; at $\alpha_\text{interp}$=1, the original representation is preserved. We refer to the former as \emph{centroid erasure}: it erases within-cluster variation while preserving cluster identity. Accuracy costs and gains are absolute changes in accuracy, reported in percent and macro-averaged across tasks unless stated otherwise. The text intervention covers the full post-image sequence, including the question, instructions, and answer options, but not pre-visual template tokens. We apply the same replacement rule separately to text and visual tokens, then measure the resulting accuracy cost on BLINK~\citep{fu2024blink}, a benchmark of visual perception tasks reformulated as multiple-choice questions.

We study six of BLINK's fourteen tasks in depth: Forensic Detection, Visual Similarity, Art Style, Counting, Relative Depth, and Spatial Relation (\Cref{app:prompts}). They provide measurable baselines and span a broad range of perceptual demands. The CD results separate them into two descriptive groups, \textsc{text-competes} and \textsc{text-needed}, which we analyze in \Cref{sec:correcting}; because these labels are assigned from the observed outcomes, they do not define an independent task taxonomy. For Qwen2.5-VL-7B, several omitted tasks have low baselines under the fixed four-letter readout, while Multi-view Reasoning and Jigsaw are higher and were not included in our exploration (\Cref{app:tasks}). Fixed-protocol results for all fourteen tasks appear in \Cref{app:blink14}. Evaluations use this fixed readout and deterministic greedy decoding; Wilson intervals and two centroid-refit sensitivity analyses are reported in \Cref{app:stats}, with model details in \Cref{app:models}.

\begin{figure}[!htbp]
    \centering
    \includegraphics[width=0.85\textwidth]{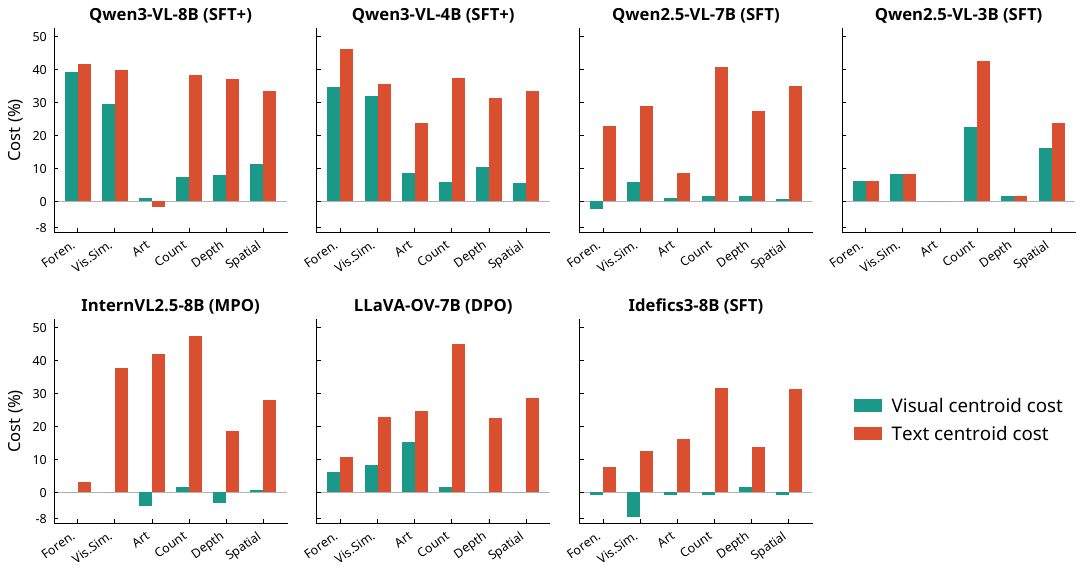}
    \vspace{-0.5em}
    \caption{\textbf{The text--visual asymmetry is consistent across models.} Per-task centroid replacement cost (visual in teal, text in red) across seven models. Text replacement is more costly for every model when averaged across tasks and in 37 of 42 model--task comparisons, with four ties and one reversal.}
    \label{fig:model_cost_grid}
\end{figure}

\subsection{Within-Cluster Visual Variation Is Largely Redundant at the Decision Layer}
\label{sec:visual_cost}

Replacing all visual tokens with their nearest centroids ($\alpha_\text{interp}$=0) costs remarkably little at the decision layer. Across seven models, the mean accuracy drop is only 6.5\% (\Cref{tab:asymmetry}). InternVL2.5-8B~\citep{chen2024expanding} and Idefics3-8B~\citep{laurencon2024building} actually improve slightly ($-$0.9\% and $-$1.5\% cost), indicating that the removed within-cluster variation can sometimes be mildly harmful under this protocol. This is consistent with prior observations that visual token compression preserves performance~\citep{chen2024image,yang2025visionzip}. The result is specific to the decision layer: an exploratory stage probe finds a 22\% cost at the vision-encoder output (\Cref{sec:mechanism}).

\subsection{Text Centroids Are Costly: The 4.0\texorpdfstring{$\times$}{x} Asymmetry}
\label{sec:text_cost}

Applying the same centroid replacement to post-image text tokens yields a dramatically different result. Collapsing these tokens to their nearest centroids costs 25.9\% accuracy on average, 4.0$\times$ the visual cost (\Cref{tab:asymmetry}). Text cost exceeds visual cost in 37 of 42 model--task comparisons, with four ties and one reversal (\Cref{fig:model_cost_grid}). The two interventions share the fitting procedure, number of centroids, images, and evaluation, but use different layers (text L12, visual L16) and token counts. The asymmetry therefore measures dependence at these two intervention points rather than a matched-layer causal effect. Under this protocol, predictions depend more on fine-grained post-image text structure than on fine-grained visual structure. Because the replaced text includes the question, instructions, and answer options, the probe does not separate harmful language priors from necessary task-interface text; distinguishing the two would require matched-layer and matched-damage controls.

\begin{table}[!htbp]
\centering
\caption{\textbf{Centroid replacement cost across seven models.} Mean accuracy cost when replacing visual or post-image text tokens with their nearest centroids ($\alpha_\text{interp}$=0). Text replacement costs 4.0$\times$ more than visual on average. Negative values indicate that replacement \emph{improves} accuracy. $\dagger$\,Ratio uses $|$vis.\ cost$|$; per-task breakdown in Figure~\ref{fig:model_cost_grid}.}
\label{tab:asymmetry}
\footnotesize
\begin{tabular}{l c c r r r}
\toprule
\textbf{Model} & \textbf{Family} & \textbf{Training} & \textbf{Vis. Cost} & \textbf{Text Cost} & \textbf{Ratio} \\
\midrule
Qwen3-VL-8B & Qwen~\citep{bai2025qwen3} & SFT+ & 16.1\% & 31.5\% & 2.0$\times$ \\
Qwen3-VL-4B & Qwen~\citep{bai2025qwen3} & SFT+ & 16.2\% & 34.7\% & 2.1$\times$ \\
Qwen2.5-VL-7B & Qwen~\citep{bai2025qwen25} & SFT & 1.4\% & 27.2\% & 19.2$\times$ \\
Qwen2.5-VL-3B & Qwen~\citep{bai2025qwen25} & SFT & 9.1\% & 13.7\% & 1.5$\times$ \\
InternVL2.5-8B & InternVL~\citep{chen2024expanding} & MPO & $-$0.9\% & 29.4\% & 34.4$\times$\textsuperscript{$\dagger$} \\
LLaVA-OV-7B & LLaVA~\citep{li2024llava} & DPO & 5.2\% & 25.8\% & 4.9$\times$ \\
Idefics3-8B & Idefics~\citep{laurencon2024building} & SFT & $-$1.5\% & 18.9\% & 12.7$\times$\textsuperscript{$\dagger$} \\
\midrule
\textbf{Mean} & & & \textbf{6.5\%} & \textbf{25.9\%} & \textbf{4.0$\times$} \\
\bottomrule
\end{tabular}
\end{table}

\subsection{Consistency Across Architectures}
\label{sec:consistency}

The text--visual asymmetry is consistent across all seven models spanning four architecture families (\Cref{fig:model_cost_grid}, \Cref{tab:asymmetry}). The ratio of text-to-visual cost ranges from 1.5$\times$ (Qwen2.5-VL-3B) to 34.4$\times$ (InternVL2.5-8B); for every model, text replacement is more costly when averaged across the six tasks. The magnitude varies in informative ways. SFT/DPO-labeled models show visual costs from $-$2 to 9\% and text costs from 14 to 27\%, while the two Qwen3 SFT+ checkpoints show elevated costs for both modalities (16\% visual, 31--35\% text). InternVL2.5-8B and Idefics3-8B, the two models with negative visual cost, show the lowest decision-layer sensitivity to within-cluster visual variation. These patterns show that the asymmetry is not confined to one architecture or training recipe, although architecture, data, tokenization, baseline accuracy, and training vary together and prevent attributing the differences to any one factor.

\paragraph{The signal is robust to norm-tail filtering.} To test whether attention-sink or near-zero-norm tokens distort the centroid fit~\citep{kang2025see, luo2025sink}, we refit Qwen2.5-VL-7B under three exclusion variants. Text centroid cost remains $+27.2\%$ across all four conditions (unfiltered plus three variants), while mean best-$\alpha_\text{interp}$ CD changes by at most 0.68\% (\Cref{app:sinks}, \Cref{tab:sink_dead}). Thus, the Qwen result is not driven by either norm tail.

\subsection{Cross-Benchmark Consistency}
\label{sec:cross_benchmark}

The asymmetry is not confined to BLINK. Across VPBench, MedBLINK, and MME, the dependence \emph{measurement} generalizes more broadly than TCCD recovery.

\paragraph{Measurement.} Text versus visual replacement costs are +11.4\% versus +3.7\% on VPBench (3.1$\times$ across seven models) and +9.3\% versus +4.6\% on MedBLINK (2.0$\times$ across nine models, including two MedGemma checkpoints~\citealp{sellergren2025medgemma,sellergren2026medgemma15}). VPBench's documented marker sensitivity remains an untested input-level confound~\citep{vptblink2026fragile}. On balanced, option-letter-free MME, text cost is at least visual cost for all ten models with validated visual-token spans (greater on nine, with one tie). For Qwen2.5-VL-7B, text replacement lowers balanced accuracy from 85.7\% to 50.0\%, while visual replacement leaves it at 85.8\%. This excludes an A/B/C/D answer-letter space and a fixed yes bias as sufficient explanations, although text replacement may still induce a default to ``no'' or disrupt task-critical predicate text. A broader 10-model $\times$ 8-benchmark MCQA grid uses a different answer-token readout and is therefore reported separately as a scoring sensitivity (\Cref{app:grid_detail}). Full portfolio and MME results appear in \Cref{app:crossbench}.

\paragraph{Recovery.} Recovery is narrower than the dependence measurement. Under the primary answer-token readout, non-oracle gains occur mainly on BLINK (5 of 7 models under cross-validated $\alpha_\text{interp}$). Auxiliary forced-choice analyses are mixed on MMBench and null on MMMU and MathVista. On free-form tasks, first-token TCCD hurts OK-VQA, centroid-replaced captioning reduces object recall, and both DocVQA replacements collapse ANLS. Complete protocols and results appear in \Cref{app:crossbench,app:cd_baselines,app:generative}.

\section{Correcting Modal Competition at Inference Time}
\label{sec:correcting}

The 4$\times$ text--visual asymmetry from \Cref{sec:measuring} suggests that a text-replaced pass may provide a useful contrastive reference. We test this idea with text centroid contrastive decoding (TCCD), which can recover visual perception accuracy on some tasks and models without retraining.

\subsection{Text Centroid Contrastive Decoding}
\label{sec:text_cd}

\begin{wraptable}[16]{r}{0.55\textwidth}
\vspace{-1.0em}
\centering
\caption{\textbf{Text centroid CD results on Qwen2.5-VL-7B.} Per-task accuracy and change under oracle best per-task $\alpha_\text{interp}$, with the fixed protocol ($\alpha_\text{interp}$=0.4) shown separately. The best-per-task column is an oracle upper bound; all significance tests use the fixed protocol.}
\label{tab:cd_results}
\footnotesize
\setlength{\tabcolsep}{3pt}
\begin{tabular}{l c c c c c}
\toprule
\textbf{Task} & $n$ & \textbf{Base} & \textbf{+CD} & $\Delta$ best & $\Delta$ 0.4 \\
\midrule
Forensic & 132 & 47.7\% & 59.1\% & \textcolor{ForestGreen}{+11.4\%} & \textcolor{ForestGreen}{+7.6\%} \\
Vis.\ Sim. & 135 & 76.3\% & 86.7\% & \textcolor{ForestGreen}{+10.4\%} & \textcolor{ForestGreen}{+8.9\%} \\
Art Style & 117 & 55.6\% & 62.4\% & \textcolor{ForestGreen}{+6.8\%} & \textcolor{ForestGreen}{+6.8\%} \\
Counting & 120 & 66.7\% & 69.2\% & \textcolor{ForestGreen}{+2.5\%} & \textcolor{ForestGreen}{+0.8\%} \\
Depth & 124 & 79.0\% & 80.6\% & \textcolor{ForestGreen}{+1.6\%} & \textcolor{red}{$-$2.4\%} \\
Spatial & 143 & 88.8\% & 88.8\% & +0\% & \textcolor{red}{$-$2.1\%} \\
\midrule
\textbf{Mean} & & 69.0\% & 74.5\% & \textbf{\textcolor{ForestGreen}{+5.4\%}} & \textbf{\textcolor{ForestGreen}{+3.3\%}} \\
\bottomrule
\end{tabular}
\vspace{-0.75em}
\end{wraptable}

Our method proceeds in two steps. First, we erase within-cluster residual structure at layer L12 by replacing each post-image text token with its nearest centroid, interpolated by a strength parameter $\alpha_\text{interp}$ (\Cref{sec:measuring}). Second, we run both the original and replaced forward passes and contrastively decode, $\text{logits}_\text{cd} = \text{logits}_\text{orig} + \alpha_\text{cd} \cdot (\text{logits}_\text{orig} - \text{logits}_\text{replaced})$, where $\alpha_\text{cd}$ controls the contrastive strength. Throughout the primary protocol, we fix $\alpha_\text{cd}$=1, as in standard VCD, LCD, DoLa, and SDCD, and tune only $\alpha_\text{interp}$. Our fixed protocol uses $\alpha_\text{interp}$=0.4 (\Cref{app:fixed_alpha}); \Cref{app:alpha} reports the full seven-model sweep and separates leave-one-task-out and oracle best-per-task settings from the exploratory sensitivity analyses.

\Cref{tab:cd_results} shows per-task results on Qwen2.5-VL-7B. At oracle best per-task $\alpha_\text{interp}$, TCCD improves or maintains accuracy on all six tasks (mean +5.4\%; +3.3\% at the fixed setting). The largest oracle gains occur on Forensic Detection (+11.4\%), Visual Similarity (+10.4\%), and Art Style (+6.8\%); we call these \textsc{text-competes} tasks. Counting (+2.5\%), Spatial Relation (+0.0\%), and Relative Depth (+1.6\%) form the \textsc{text-needed} group. These labels and the oracle gains are defined on the same outcomes; the corresponding per-task tests are therefore nominal, while fixed-setting inference is reported in \Cref{app:fixed_alpha}.

\begin{wrapfigure}{r}{0.50\textwidth}
\vspace{-1.5em}
\centering
\includegraphics[width=0.48\textwidth]{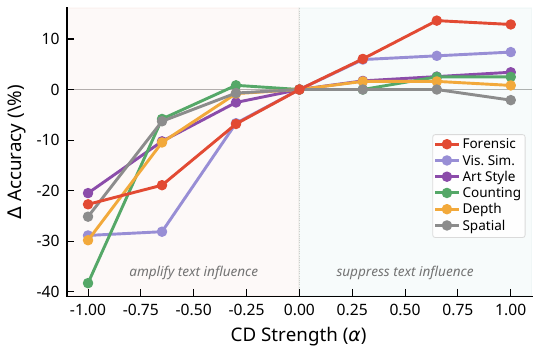}
\vspace{-0.5em}
\caption{\textbf{Directional, task-selective dose--response} (Qwen2.5-VL-7B, $\alpha_\text{interp}$=0.4; exploratory 11{,}218-image held-out COCO+ScienceQA cache, $K$=512). Negative $\alpha_\text{cd}$ hurts every task; positive $\alpha_\text{cd}$ selectively benefits the post-hoc \textsc{text-competes} group. On the negative-$\alpha_\text{cd}$ side, Counting is the sole monotonicity exception.}
\label{fig:dose_response}
\vspace{-1.0em}
\end{wrapfigure}

\paragraph{Dose--response.} To probe directionality, we sweep $\alpha_\text{cd}$ from $-$1.0 (amplifying the replaced-text reference) to +1.0 (contrasting against it and effectively suppressing it; \Cref{fig:dose_response}). Negative $\alpha_\text{cd}$ hurts every task by 20.5--38.3\% at $-$1.0 and monotonically on five of six (Counting deviates by one item at $-$0.3); positive $\alpha_\text{cd}$ selectively benefits the \textsc{text-competes} group and leaves \textsc{text-needed} near baseline. This directional, task-selective pattern is consistent with a structured signal rather than random noise, although the exploratory sweep does not by itself establish a causal mechanism.

\paragraph{Answer-logit resampling control.} Repeated answer-logit sampling does not recover the same gains. We also draw option letters repeatedly from one frozen forward-pass softmax under a seed-pinned protocol (\Cref{app:votek}). Mixed voting (one greedy draw plus $k-1$ stochastic draws at $T$=0.7, ties toward greedy) stays near the greedy result (0.0/$-$0.1/$-$0.2\% at Vote@\{2,5,10\}); pure-stochastic voting changes the mean by $-$3.2/$-$3.4/$-$0.7\%, and repeated greedy draws are identical. Unlike independent-path self-consistency, this approximately 1$\times$-compute control isolates sampling variance without generating separate reasoning paths~\citep{wang2022self}. TCCD uses two forward passes and gains +3.3\% at the fixed protocol (+5.4\% oracle).

\begin{wraptable}[15]{r}{0.53\textwidth}
\vspace{-1.0em}
\centering
\caption{\textbf{Fixed-setting contrastive decoding baselines} (Qwen2.5-VL-7B, six BLINK tasks, $n$=771). Mean accuracy change under the primary bare-continuation readout at $\alpha_\text{cd}$=1; TCCD also fixes $\alpha_\text{interp}$=0.4. Methods use separate runs on the same task rows.}
\label{tab:cd_baselines}
\footnotesize
\setlength{\tabcolsep}{2.6pt}
\begin{tabular}{l l r}
\toprule
\textbf{Method} & \textbf{Protocol} & \textbf{Mean $\Delta$} \\
\midrule
Text centroid CD (ours) & fixed & \textcolor{ForestGreen}{+3.27\%} \\
Text-only LCD & fixed & \textcolor{ForestGreen}{+2.43\%} \\
VCD & fixed & \textcolor{ForestGreen}{+2.03\%} \\
DoLa low / high & fixed & \textcolor{ForestGreen}{+1.61\% / +3.42\%} \\
SDCD & fixed & \textcolor{ForestGreen}{+0.87\%} \\
\bottomrule
\end{tabular}
\vspace{-0.75em}
\end{wraptable}

\paragraph{Comparison to contrastive decoding baselines.} \Cref{tab:cd_baselines} compares fixed-setting runs on the same six BLINK tasks and primary answer-token readout. TCCD (+3.27\%) exceeds text-only LCD (+2.43\%), VCD (+2.03\%), DoLa-low (+1.61\%), and SDCD (+0.87\%), while DoLa-high is marginally higher (+3.42\% versus +3.27\%, a difference of 0.15 percentage points). All methods use the same task rows, scoring, and contrastive strength, but come from separate runs; TCCD is competitive rather than uniformly best. On a separate primary-scored blank-image LCD run, TCCD and LCD make complementary errors, with 36 and 37 uniquely correct items, respectively. A label-free majority vote improves the better method by +1.3\%, compared with a +4.7\% label-dependent oracle union. Further baseline, scoring, MMBench, visual-side, and external-judge analyses appear in \Cref{app:cd_baselines,app:visccd,app:gemini}.

\paragraph{Practical operating points.} At a shared $\alpha_\text{interp}=0.4$, Qwen2.5-VL-7B gains +3.3\% on average. Pooling paired predictions across the six tasks, TCCD corrects 75 baseline errors while changing 50 correct answers to errors (nominal two-sided McNemar $p=0.025$; asymptotic, without continuity correction). No individual task is significant after Bonferroni correction for six comparisons. In an exploratory in-sample policy sweep, applying the second forward pass to the 50\% least-confident answers by top-2 logit margin retains +3.5\% at 1.5$\times$ compute (blanket CD in the same experiment: +3.2\% pooled at 2$\times$); the selection rule and coverage require held-out calibration. Across models, leave-one-task-out CV-$\alpha_\text{interp}$ yields a modest +0.9\% mean, positive on 5 of 7 models and at or below zero on two (Qwen3-VL-4B $-$5.8\%, InternVL2.5-8B $-$1.8\%; \Cref{tab:mci}).

\begin{table}[!htbp]
\centering
\caption{\textbf{TCCD effect across reported training approaches.} Six-task mean baseline accuracy and TCCD change under fixed $\alpha_\text{interp}$=0.4, leave-one-task-out selection, and oracle best-per-task selection. Here $g=(a_\text{TCCD}-a_\text{base})/(1-a_\text{base})$, averaged across the six tasks. Per-(model, task) Wilson CIs appear in \Cref{app:alpha}.}
\label{tab:mci}
\footnotesize
\begin{tabular}{l c r r r r r}
\toprule
\textbf{Model} & \textbf{Train.} & \textbf{Base} & $\Delta$ $\alpha_\text{interp}$=0.4 & $\Delta$ CV-$\alpha_\text{interp}$ & $\Delta$ best-$\alpha_\text{interp}$ & $g$ \\
\midrule
Qwen3-VL-8B & SFT+ & 73\% & \textcolor{BrickRed}{$-$2.1\%} & \textcolor{ForestGreen}{+0.9\%} & \textcolor{ForestGreen}{+1.6\%} & +0.04 \\
Qwen3-VL-4B & SFT+ & 76\% & \textcolor{BrickRed}{$-$6.8\%} & \textcolor{BrickRed}{$-$5.8\%} & \textcolor{BrickRed}{$-$0.2\%} & $-$0.01 \\
Qwen2.5-VL-7B & SFT & 69\% & \textcolor{ForestGreen}{+3.3\%} & \textcolor{ForestGreen}{+3.3\%} & \textcolor{ForestGreen}{+5.4\%} & +0.16 \\
Qwen2.5-VL-3B & SFT & 55\% & \textcolor{ForestGreen}{+3.9\%} & \textcolor{ForestGreen}{+5.9\%} & \textcolor{ForestGreen}{+8.6\%} & +0.19 \\
InternVL2.5-8B & MPO & 68\% & \textcolor{BrickRed}{$-$1.0\%} & \textcolor{BrickRed}{$-$1.8\%} & \textcolor{ForestGreen}{+3.1\%} & +0.11 \\
LLaVA-OV-7B & DPO & 68\% & \textcolor{ForestGreen}{+2.0\%} & \textcolor{ForestGreen}{+1.5\%} & \textcolor{ForestGreen}{+4.7\%} & +0.14 \\
Idefics3-8B & SFT & 59\% & \textcolor{ForestGreen}{+0.5\%} & \textcolor{ForestGreen}{+2.4\%} & \textcolor{ForestGreen}{+3.8\%} & +0.09 \\
\midrule
\textit{SFT/DPO avg.} & & & \textcolor{ForestGreen}{+2.4\%} & \textcolor{ForestGreen}{+3.3\%} & \textbf{\textcolor{ForestGreen}{+5.6\%}} & +0.15 \\
\textit{SFT+/MPO avg.} & & & \textcolor{BrickRed}{$-$3.3\%} & \textcolor{BrickRed}{$-$2.2\%} & \textbf{\textcolor{ForestGreen}{+1.5\%}} & +0.05 \\
\bottomrule
\end{tabular}
\end{table}

\subsection{Where Is the Intervention Effective?}
\label{sec:mechanism}

\paragraph{Positional-span ablation.} We selectively replace four positional ranges: all post-image text, its first 70\% or last 30\%, and the pre-visual prefix (\Cref{app:segdose}). These are not parsed question/option segments, and the pre-visual prefix includes template and control tokens. At $\alpha_\text{interp}=0.4$, all-post-image replacement is strongest overall (Forensic $+7.6\%$, Vis.\ Sim. $+8.9\%$, Art Style $+6.8\%$), while the pre-visual prefix gives Forensic $+12.1\%$. Across doses, the pre-visual prefix is the only single range with a positive \textsc{text-competes} mean throughout, the post-image suffix helps only at reduced doses, and no single range matches the full post-image span's +7.8\% optimum. The ablation maps where the intervention is effective without assigning a semantic or causal locus.

\paragraph{Layer sweep.} We apply text centroid CD at 16 individual layers (L0--L26) of Qwen2.5-VL-7B (\Cref{fig:layer_sweep}, \Cref{app:layers}). The \textsc{text-competes} group benefits consistently from L4 through L22, while \textsc{text-needed} is hurt over much of the same range; effects diminish near the extremes. Thus, the pattern is not an artifact of choosing L12, although the groups themselves remain outcome-defined.

\paragraph{Where visual representations become coarsely summarizable.} An exploratory Qwen2.5-VL-7B probe finds a visual cost of 22\% at the vision-encoder output, 21\% after the projector, and 2\% at L16; a layer sweep places the drop between L12 and L16 ($n$=771; \Cref{app:stage_probe}). This is consistent with visual states becoming centroid-summarizable by the middle of the LLM and provides visual-side support independent of the text task interface. Stage-specific fits use partly different samples, so the localization remains exploratory.

\paragraph{Specificity control.} To test whether the learned centroid directions matter, we compare TCCD with exactly equal-norm random directions. At the fixed primary protocol ($\alpha_\text{interp}=0.4$), nearest-centroid TCCD gains $+7.77\%$ on the \textsc{text-competes} mean versus $+0.24\%$ for the random directions; across all six tasks, the means are $+3.27\%$ versus $+0.40\%$ (seed 42; \Cref{app:controls}). This single-seed comparison supports sensitivity to the tested centroid direction beyond displacement magnitude alone, but one random-direction null does not establish that $K$-means is unique.

\paragraph{Relation to distribution-shift explanations.} Any contrastive reference can shift the output distribution~\citep{yin2025mirage}. We therefore compare TCCD with equal-norm random directions and sweep $\alpha_\text{cd}$ in both directions. Their different behavior suggests that the gains are not explained by perturbation magnitude alone, but these controls do not rule out every distribution-shift explanation or identify a unique mechanism.

\section{Quantifying Modal Dependence and Correctability}
\label{sec:diagnostic}

TCCD does not uniformly improve every model or task (\Cref{sec:correcting}). We therefore distinguish two complementary quantities. The primary signal is the text--visual accuracy-cost asymmetry, whose aggregate ordering also recurs on VPBench, MedBLINK, and MME (\Cref{sec:cross_benchmark}). TCCD recovery is a secondary measure of \emph{correctability}: how much performance is recoverable under this particular intervention. It varies with $\alpha_\text{interp}$, checkpoint, baseline accuracy, and task composition. Together, the two quantities describe both modal dependence and how much of the measured dependence is actionable without retraining.

\subsection{Correctability Varies Across Checkpoints}
\label{sec:spectrum}

\Cref{tab:mci} separates recovery under a shared fixed setting, task-held-out calibration, and an oracle upper bound. TCCD improves 4 of 7 checkpoint means at fixed $\alpha_\text{interp}=0.4$, 5 of 7 under leave-one-task-out selection, and 6 of 7 under oracle best-per-task selection. The oracle results show that the intervention exposes recoverable performance, while their gap from the non-oracle results shows that selecting its strength remains part of the problem. In this seven-checkpoint sample, the four SFT/DPO-labeled checkpoints average +5.6\% under oracle selection, compared with +1.5\% for the three SFT+/MPO-labeled checkpoints. Fixed (+2.4\% vs.\ $-$3.3\%) and leave-one-task-out (+3.3\% vs.\ $-$2.2\%) settings preserve this descriptive ordering, although recovery remains negative for some individual checkpoints.

Some of this variation tracks baseline headroom. Across the 42 oracle-selected model--task cells, a crossed restricted maximum-likelihood model with model and task random intercepts estimates a 0.213-percentage-point decrease in recovery for each one-percentage-point increase in baseline accuracy (SE 0.052; nominal Wald $p=4.57\times10^{-5}$; marginal $R^2=0.379$). Yet after normalizing each gain by the remaining error, Hake gain is positive for 6 of 7 model averages (mean $g=+0.103$; nominal two-sided one-sample $t$-test $p=0.0076$). Thus, under oracle tuning, TCCD recovers about 10.3\% of the remaining errors on average in this sample. Recovery therefore remains positive after accounting for available headroom, so baseline performance and intervention-based correctability provide complementary summaries of a checkpoint.

These analyses do not isolate a training-recipe effect. The reported training labels covary with architecture, model family, data, and baseline performance; recovery already subtracts baseline accuracy; and both statistical summaries use oracle-selected outcomes. A controlled comparison would vary training while holding the base model and data fixed. Establishing whether TCCD correctability tracks independent gradient- or attention-based measures of modal balance also remains future work.

We therefore interpret TCCD recovery as correctability under the tested intervention, not as a direct measure of whether training has balanced the modalities. A large gain shows that some errors are recoverable without retraining under that protocol; a near-zero gain can instead reflect ceiling effects, task composition, or architecture-specific sensitivity. The primary audit signal remains the text--visual accuracy-cost asymmetry, while TCCD response adds a secondary view of how much measured dependence is actionable. Together, they provide a lightweight labeled audit that complements training-time diagnostics~\citep{wu2025mitigating}: the audit requires fitted centroid banks, labels, and one replacement pass per modality, but no retraining.

\section{Discussion and Conclusion}
\label{sec:discussion}

\paragraph{Broader implications.} Our findings suggest that the 4$\times$ text--visual asymmetry is not peculiar to one architecture: the aggregate ordering recurs across the evaluated model families and perception-focused benchmarks. This has practical consequences. In perception-critical settings such as medical imaging~\citep{asadi2026mirage} and agent pipelines~\citep{heydari2025anatomy}, centroid replacement offers a lightweight way to measure dependence on fine-grained text versus visual structure using a labeled audit set. For model developers, it complements ordinary accuracy at the aggregate evaluation level, although it does not certify that any individual answer is visually grounded. The decision-layer coarseness observed in \Cref{sec:visual_cost} is also consistent with geometric explanations for visual token compression~\citep{chen2024image,yang2025visionzip}.

\paragraph{Application to monitoring.} Our probe maps onto the functionality-monitoring gap identified by NIST AI 800-4~\citep{rao2026challenges}. A concrete example is Qwen2.5-VL-7B on VPBench depth estimation: it reaches 64.7\% accuracy while its measured visual cost is $-0.7\%$ and text cost is $+14.4\%$ (aggregate $+0.6\%$/$+11.0\%$, \Cref{tab:vpbench_summary}). Task-level accuracy does not surface this dissociation; centroid replacement does. Operational use would require a labeled sentinel set and periodic re-measurement.

\paragraph{Limitations.} We evaluate the primary six-task analysis on BLINK validation because test-set answers are not public~\citep{fu2024blink,fu2025hiddenplainsightvlms}. The task grouping is exploratory, and tests after oracle $\alpha_\text{interp}$ selection are nominal. Across five Qwen2.5-VL-7B centroid refits, the maximum per-task standard deviation at the primary fit's frozen oracle $\alpha_\text{interp}$ is 2.5 percentage points. On Qwen2.5-VL-7B, fixed-dose TCCD raises aggregate ECE by +0.028, reduced to +0.006 under an exploratory selective-coverage policy. Demonstrated TCCD recovery is limited to discriminative, predominantly forced-choice visual QA; our generative and OCR experiments do not show a benefit. The text and visual probes use L12 and L16 with unequal token counts, without matching either the information removed or the damage induced. Post-image text also contains the task interface, so full replacement is a dependence stress test rather than evidence of a harmful language prior; the fixed TCCD protocol uses partial replacement. A fully protocol-matched same-layer comparison and a cross-modal control matched for task-relevant information loss remain future work.

\paragraph{Conclusion.} Centroid replacement reveals a 4$\times$ text--visual cost asymmetry across seven multimodal language models, with the same aggregate ordering on VPBench, MedBLINK, and a balanced, option-letter-free MME control. TCCD exploits this structure to recover up to +16.9\% accuracy on an individual BLINK task without retraining, at oracle best-per-task $\alpha_\text{interp}$; recovery varies across tasks and models under fixed and cross-validated settings. We frame the asymmetry as a labeled observational audit rather than a causal training signal or label-free monitor. An exploratory stage probe further traces visual cost from 22\% at the encoder to 2\% by L16. Together, centroid replacement exposes a recurring modal imbalance and turns part of it into a reusable, retraining-free intervention.

\section{Acknowledgements}
\label{sec:acknowledgements}

This work was supported primarily by the European Research Council under the European Union's Horizon 2020 research and innovation program (Grant 804226, PERDY), with additional support from the U.S. National Institutes of Health (Grant AG08916) and a Stanford Institute for Human-Centered Artificial Intelligence (HAI) Hoffman-Yee Award.

\clearpage
\bibliography{main}
\bibliographystyle{colm2026_conference}

\clearpage

\appendix
\section{Evaluation Details}
\label{app:eval}

\subsection{Models and Implementation}
\label{app:models}
We evaluate seven instruction-tuned MLMs spanning four architecture families (Qwen~\citep{bai2025qwen3}, InternVL~\citep{chen2024expanding}, LLaVA~\citep{li2024llava}, Idefics~\citep{laurencon2024building}) and four training-recipe labels reported for the checkpoints (SFT, DPO~\citep{rafailov2023direct}, SFT+, MPO~\citep{zhu2025internvl3}). Table~\ref{tab:supp_models} lists all models with their HuggingFace identifiers and vision encoders. All models are loaded in bfloat16 precision on a single NVIDIA A6000 (48GB). Unless otherwise noted, model evaluations use greedy decoding, eliminating sampling variance, although low-level hardware and kernel nondeterminism may remain.

\paragraph{Why Qwen2.5-VL-7B is the primary deep-dive model.} We center our detailed analyses (Figures~\ref{fig:dose_response},~\ref{fig:segment_ablation},~\ref{fig:layer_sweep} and Tables~\ref{tab:cd_results},~\ref{tab:selfconsistency}) on Qwen2.5-VL-7B because it is a widely benchmarked open MLM at a practical 7B scale~\citep{fu2024blink,fu2025hiddenplainsightvlms,deng2025words,qi2025beyond} and its response is sufficiently large for detailed geometric analysis. We evaluate the core dependence measurement and TCCD response across six additional checkpoints spanning three other architecture families. These cross-sectional comparisons establish breadth but do not isolate training-recipe effects.

\begin{table}[!htbp]
\centering
\caption{\textbf{Model registry.} Seven instruction-tuned MLMs spanning four architecture families, four reported training-recipe labels, and parameter scales from 3B to 8B. All use HuggingFace checkpoints.}
\vspace{1em}
\label{tab:supp_models}
\small
\resizebox{\textwidth}{!}{%
\begin{tabular}{l l l l l l}
\toprule
\textbf{Model} & \textbf{Family} & \textbf{Train.} & \textbf{Size} & \textbf{HuggingFace ID} & \textbf{Vis.\ Enc.} \\
\midrule
Qwen2.5-VL-3B & Qwen & SFT & 3B & \texttt{Qwen/Qwen2.5-VL-3B-Instruct} & Qwen2-VL \\
Qwen2.5-VL-7B & Qwen & SFT & 7B & \texttt{Qwen/Qwen2.5-VL-7B-Instruct} & Qwen2-VL \\
Qwen3-VL-4B & Qwen & SFT+ & 4B & \texttt{Qwen/Qwen3-VL-4B-Instruct} & Qwen3-VL \\
Qwen3-VL-8B & Qwen & SFT+ & 8B & \texttt{Qwen/Qwen3-VL-8B-Instruct} & Qwen3-VL \\
InternVL2.5-8B & InternVL & MPO & 8B & \texttt{OpenGVLab/InternVL2\_5-8B-MPO-hf} & InternViT \\
LLaVA-OV-7B & LLaVA & DPO & 7B & \texttt{llava-hf/llava-onevision-qwen2-7b-ov-hf} & SigLIP \\
Idefics3-8B & Idefics & SFT & 8B & \texttt{HuggingFaceM4/Idefics3-8B-Llama3} & SigLIP \\
\bottomrule
\end{tabular}}
\end{table}

\subsection{Centroid Fitting Procedure}
\label{app:centroids}

We fit $K$-means centroids ($K$=256) on hidden-state activations extracted from layer L12 (text) and L16 (visual, used for \Cref{sec:visual_cost}). For the cross-model comparison (Tables~\ref{tab:asymmetry} and~\ref{tab:mci}) we use 2{,}000 MS-COCO images streamed with data seed 1337, $K$-means seed 42, and $\alpha_\text{cd}$=1.0. Each image is rendered with the fixed harvest prompt ``Describe what you see in this image.'', followed by a newline and ``Answer:''; the text bank retains 16 post-image tokens per image (32{,}000 total). The fit uses no BLINK labels or evaluation prompts. Its generic prompt does not cover BLINK's diverse semantics, and possible image-source overlap was not measured. This is the \emph{primary protocol}; earlier exploratory analyses use an 11{,}218-image held-out COCO+ScienceQA cache with $K$=512. Visual and text tokens are separated by position and fitted independently. A 30-cell grid spans $N\in[1{,}000,50{,}000]$ and $K\in[128,2{,}048]$ (mean oracle-best $\Delta=5.2\%\pm0.4\%$ across cells; \Cref{app:ksweep}); $N$=2{,}000, $K$=256 is a low-compute setting near the lower end of that grid, not its minimum. The modality fits differ substantially in token count (707{,}414 visual vs.\ 32{,}000 text tokens before filtering; \Cref{tab:sink_dead}), so the comparison uses parallel rather than token-count-matched fits.

\clearpage

\subsection{Benchmark and Prompt Format}
\label{app:prompts}
\begin{figure}[!htbp]
    \centering
    \includegraphics[width=\textwidth]{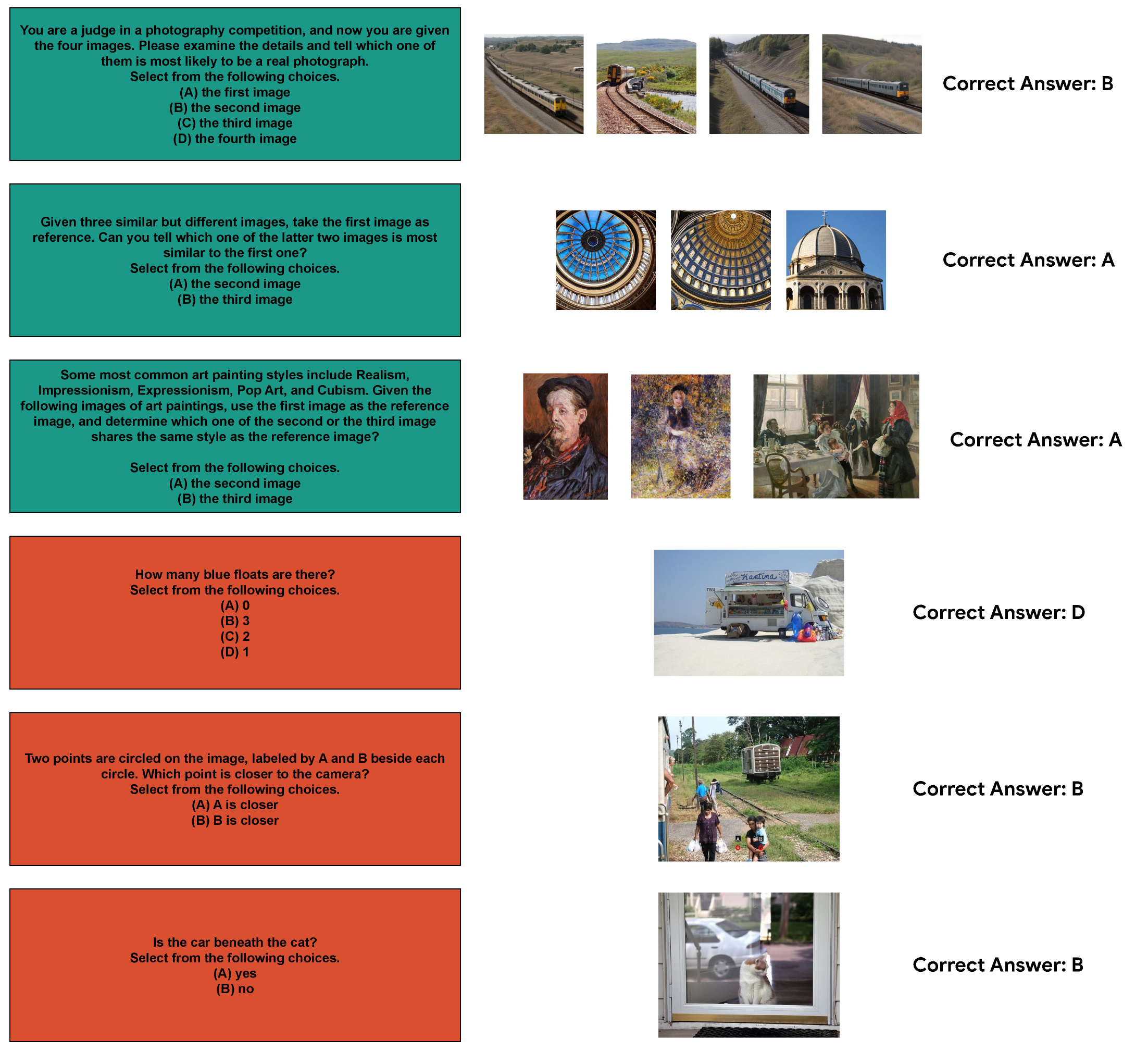}
    \caption{\textbf{Task examples from BLINK.} One representative sample per task showing the input image(s), prompt with answer choices, and the correct answer. Top row: \textsc{text-competes} tasks (Forensic Detection, Visual Similarity, Art Style), where TCCD yields larger gains. Bottom row: \textsc{text-needed} tasks (Counting, Relative Depth, Spatial Relation), where gains are smaller or absent.}
    \label{fig:supp_task_examples}
\end{figure}

We evaluate on BLINK~\citep{fu2024blink}, a benchmark of 14 classic computer vision tasks reformulated as visual multiple-choice questions. Each sample consists of one or more input images, a text prompt with answer options, and a single correct answer. Figure~\ref{fig:supp_task_examples} shows one representative example per task in the six-task analysis. We use a uniform four-letter answer-token readout across BLINK: at the final prompt position, we take the argmax over the bare continuation logits for A--D, the same position at which the contrastive-decoding logits are formed. Some BLINK prompts expose only two or three choices; on these items, selecting an undisplayed letter counts as an incorrect answer. The same scorer is applied to the baseline and every intervention pass, so all reported costs and gains are paired changes under a fixed readout (see the full template below).

\paragraph{Full prompt for the Figure~\ref{fig:overview}(a) sample.} We pass BLINK's question and answer options verbatim through the model's chat template, using the default Qwen2.5-VL system prompt. The resulting input for the Forensic Detection sample in Figure~\ref{fig:overview}(a) (and identically for all Forensic Detection items) is:

\begin{quote}
\small
\textbf{System:} \texttt{You are a helpful assistant.}

\textbf{User:} $\langle$image$\rangle$ \texttt{You are a judge in a photography competition, and now you are given the four images. Please examine the details and tell which one of them is most likely to be a real photograph.} \\
\texttt{Select from the following choices.} \\
\texttt{(A) the first image} \\
\texttt{(B) the second image} \\
\texttt{(C) the third image} \\
\texttt{(D) the fourth image}
\end{quote}

The four candidate images are concatenated horizontally into a single image before templating; the $\langle$image$\rangle$ placeholder expands to one pad token per visual token (1{,}314 for this sample). The answer is scored by the argmax over the option-letter logits at the final position, not by free-form generation.

\subsection{Figure~\ref{fig:overview}(a) Exemplar and Attention Calculation}
\label{app:fig1_exemplar}

\Cref{fig:overview}a uses Qwen2.5-VL-7B on BLINK Forensic Detection item \texttt{val\_Forensic\_Detection\_18} (zero-based row 17 at the pinned revision). Its four candidate images expand to a 1{,}314-token visual span (half-open indices $[15,1329)$; post-merge grid $18\!\times\!73$). Both the original pass and the pass with all post-image text replaced ($\alpha_\text{interp}$=0) predict the incorrect answer A. TCCD contrasts their output logits at $\alpha_\text{cd}$=1 and selects the correct answer B.

To measure attention, we reconstruct attention from the answer-scoring input position after Qwen's multimodal rotary embedding and causal mask, using the Transformers 5.4.0 eager-attention formula while retaining SDPA for the model forward. At L22, mean-over-head attention mass on the visual span rises from 3.63\% in the original pass to 11.09\% in the reference. At L16/L20/L22, the respective values are 2.84/3.86/3.63\% and 4.50/9.43/11.09\% (means 3.44\% and 8.34\%). The mean rises at all three layers, although only 14/28, 15/28, and 16/28 heads increase; heads are not independent replicates.

The displayed heatmaps are softmax-normalized pre-RoPE QK scores, reshaped to $36\!\times\!36$ from the first 1{,}296 visual positions. They are qualitative views rather than the post-RoPE attention measurement reported above. The right heatmap comes from the text-replaced pass; TCCD predicts B by contrasting the logits of the two passes. This example is illustrative rather than an average result.

\clearpage

\section{Task Selection and Excluded Tasks}
\label{app:tasks}

We focus on six of BLINK's 14 tasks, chosen for baseline measurability and coverage of low-, mid-, and high-level perceptual demands. Several remaining tasks have low Qwen2.5-VL-7B baselines under the fixed four-letter readout, while Multi-view Reasoning and Jigsaw are higher; the six-task analysis is therefore not a complete benchmark estimate. We assign the \textsc{text-competes}/\textsc{text-needed} labels post hoc from the CD outcomes rather than use them for task selection. Table~\ref{tab:supp_excluded} reports every remaining baseline, and \Cref{app:blink14} gives fixed-protocol results for all fourteen tasks on two models.

\begin{table}[!htbp]
\centering
\caption{\textbf{BLINK tasks outside the six-task analysis.} Qwen2.5-VL-7B baselines under the primary fixed four-letter readout. Several tasks have low baselines, while Multi-view Reasoning and Jigsaw are higher. Results on all fourteen tasks appear in \Cref{tab:blink14}.}
\vspace{1em}
\label{tab:supp_excluded}
\small
\begin{tabular}{l r}
\toprule
\textbf{Task} & \textbf{Baseline (\%)} \\
\midrule
Multi-view Reasoning & 55.6 \\
Jigsaw Puzzle Solving & 55.3 \\
Object Localization & 45.1 \\
Visual Correspondence & 32.6 \\
Semantic Correspondence & 32.4 \\
Relative Reflectance & 32.1 \\
Functional Correspondence & 17.7 \\
IQ Test & 19.3 \\
\bottomrule
\end{tabular}
\end{table}

\section{Full Alpha Sweep Results}
\label{app:alpha}

\begin{figure}[!htbp]
    \centering
    \includegraphics[width=0.6\textwidth]{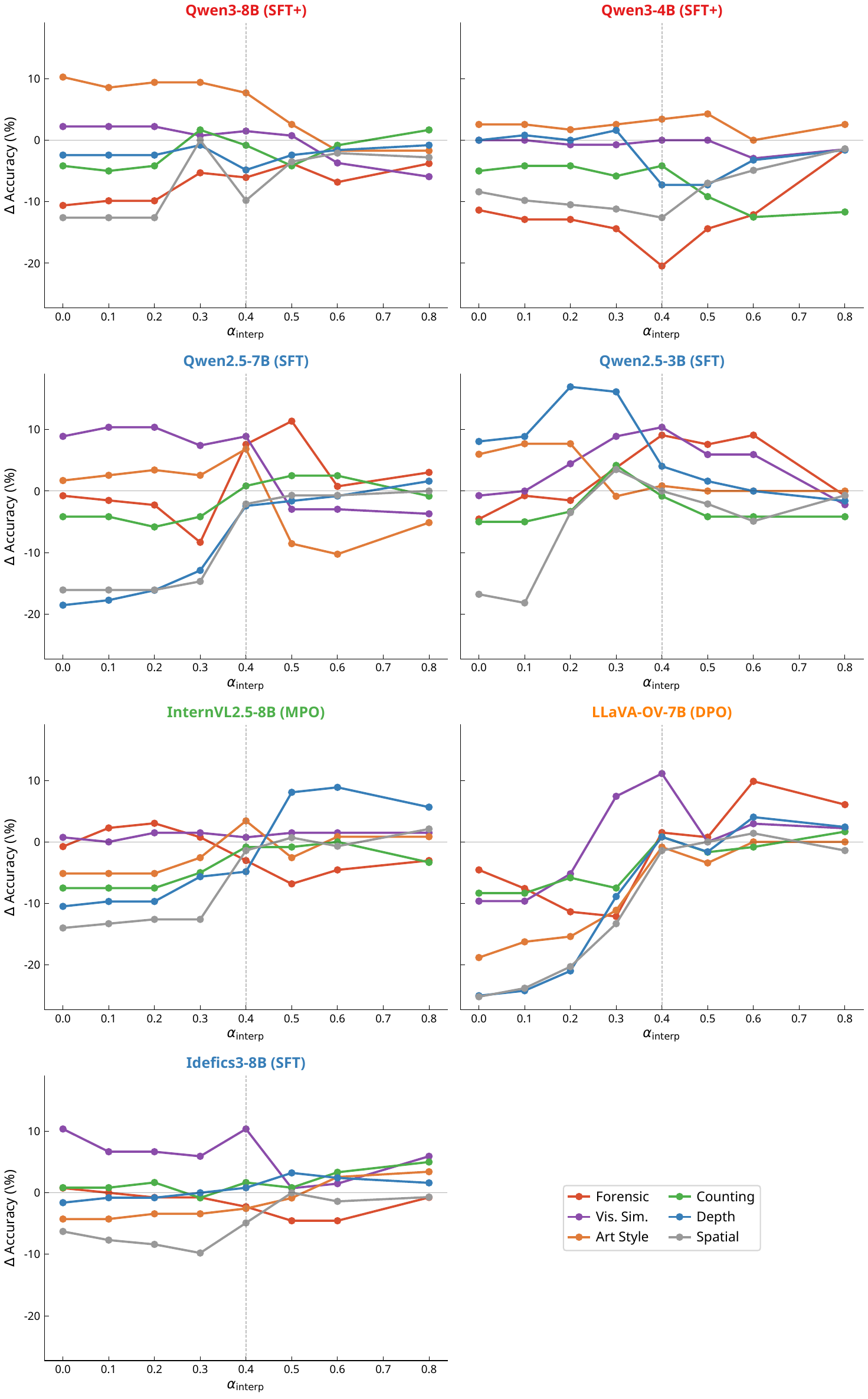}
    \caption{\textbf{Per-model alpha sweep under the primary protocol.} CD delta (\%) versus centroid interpolation strength $\alpha_\text{interp}$ for all seven models ($N$=2{,}000, $K$=256, $\alpha_\text{cd}$=1.0). The dashed vertical line marks the fixed protocol, $\alpha_\text{interp}$=0.4. SFT/DPO-labeled checkpoints (Qwen2.5, LLaVA-OV, Idefics3) show sharper task-selective peaks than SFT+/MPO checkpoints in this descriptive cross-checkpoint comparison.}
    \label{fig:supp_alpha_sweep}
\end{figure}

\begin{figure}[!htbp]
    \centering
    \includegraphics[width=0.55\textwidth]{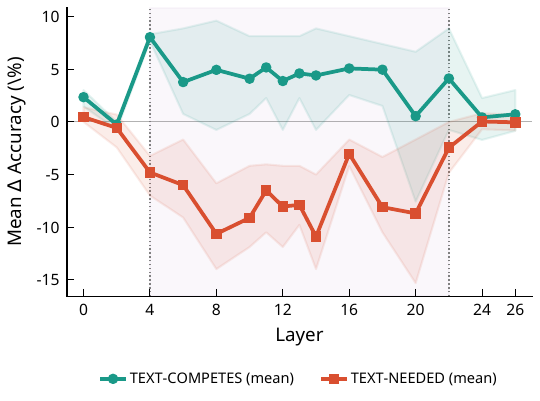}
    \caption{\textbf{Task-group separation across model depth.} Mean CD delta for \textsc{text-competes} and \textsc{text-needed} across 16 layers of Qwen2.5-VL-7B ($\alpha_\text{interp}$=0, $\alpha_\text{cd}$=0.65; exploratory 11{,}218-image held-out COCO+ScienceQA cache, $K$=512). The L12 centroid bank is injected at every layer. Group means separate from L4 through L22, indicating that the pattern is not specific to one injection layer.}
    \label{fig:layer_sweep}
\end{figure}

\Cref{fig:supp_alpha_sweep} shows the complete centroid interpolation sweep ($\alpha_\text{interp}$ from 0.0 to 0.8) for all seven models under the primary protocol ($N$=2{,}000 COCO images, $K$=256, $\alpha_\text{cd}$=1.0). Each panel displays per-task CD delta as a function of erasure strength, with the fixed protocol ($\alpha_\text{interp}$=0.4) marked by a dashed vertical line. SFT/DPO-labeled checkpoints show sharper task-selective peaks in this sample, while SFT+/MPO checkpoints are flatter. Because the checkpoints differ in other respects, this is a descriptive association rather than an isolated effect of post-training.

\subsection{Full Per-Model, Per-Task Breakdown with Wilson CIs}
\label{app:per_model_task}

\Cref{tab:per_model_task} reports the complete per-(model, task) breakdown at oracle best per-task $\alpha_\text{interp}$: baseline and TCCD accuracy with Wilson 95\% CIs, the delta, and selected $\alpha_\text{interp}$. Since $\alpha_\text{interp}$ is selected and evaluated on the same items, the table reports confidence intervals but not inferential significance; fixed-$\alpha_\text{interp}$ and CV-$\alpha_\text{interp}$ means are in \Cref{tab:mci}, and refit sensitivity is in \Cref{app:variance}.

\begin{table}[!htbp]
\centering
\caption{\textbf{Per-(model, task) TCCD at oracle best per-task $\alpha_\text{interp}$} ($N$=2{,}000 COCO images, $K$=256, $\alpha_\text{cd}$=1.0). Baseline and +CD accuracies and $\Delta$ are in \%, with Wilson 95\% CIs. Same-set selection makes this an oracle analysis.}
\vspace{1em}
\label{tab:per_model_task}
\footnotesize
\setlength{\tabcolsep}{4pt}
\begin{tabular}{l c c r c}
\toprule
\textbf{Task} & \textbf{Base [95\% CI]} & \textbf{+CD [95\% CI]} & $\Delta$ & $\alpha_\text{interp}^{*}$ \\
\midrule
\multicolumn{5}{l}{\textbf{Qwen3-VL-8B}} \\
\quad Forensic & 66.7 [58.3, 74.1] & 62.9 [54.4, 70.6] & \textcolor{red}{$-$3.8} & 0.5 \\
\quad Vis.\ Sim. & 87.4 [80.8, 92.0] & 89.6 [83.3, 93.7] & \textcolor{ForestGreen}{+2.2} & 0.0 \\
\quad Art Style & 45.3 [36.6, 54.3] & 55.6 [46.5, 64.2] & \textcolor{ForestGreen}{+10.3} & 0.0 \\
\quad Counting & 64.2 [55.3, 72.2] & 65.8 [57.0, 73.7] & \textcolor{ForestGreen}{+1.7} & 0.3 \\
\quad Depth & 88.7 [81.9, 93.2] & 87.9 [81.0, 92.5] & \textcolor{red}{$-$0.8} & 0.3 \\
\quad Spatial & 87.4 [81.0, 91.9] & 87.4 [81.0, 91.9] & +0.0 & 0.3 \\
\midrule
\multicolumn{5}{l}{\textbf{Qwen3-VL-4B}} \\
\quad Forensic & 71.2 [63.0, 78.2] & 69.7 [61.4, 76.9] & \textcolor{red}{$-$1.5} & 0.8 \\
\quad Vis.\ Sim. & 83.0 [75.7, 88.4] & 83.0 [75.7, 88.4] & +0.0 & 0.0 \\
\quad Art Style & 70.9 [62.2, 78.4] & 75.2 [66.7, 82.2] & \textcolor{ForestGreen}{+4.3} & 0.5 \\
\quad Counting & 63.3 [54.4, 71.4] & 59.2 [50.2, 67.5] & \textcolor{red}{$-$4.2} & 0.1 \\
\quad Depth & 83.1 [75.5, 88.6] & 84.7 [77.3, 90.0] & \textcolor{ForestGreen}{+1.6} & 0.3 \\
\quad Spatial & 87.4 [81.0, 91.9] & 86.0 [79.4, 90.8] & \textcolor{red}{$-$1.4} & 0.8 \\
\midrule
\multicolumn{5}{l}{\textbf{Qwen2.5-VL-7B}} \\
\quad Forensic & 47.7 [39.4, 56.2] & 59.1 [50.6, 67.1] & \textcolor{ForestGreen}{+11.4} & 0.5 \\
\quad Vis.\ Sim. & 76.3 [68.5, 82.7] & 86.7 [79.9, 91.4] & \textcolor{ForestGreen}{+10.4} & 0.1 \\
\quad Art Style & 55.6 [46.5, 64.2] & 62.4 [53.4, 70.6] & \textcolor{ForestGreen}{+6.8} & 0.4 \\
\quad Counting & 66.7 [57.8, 74.5] & 69.2 [60.4, 76.7] & \textcolor{ForestGreen}{+2.5} & 0.5 \\
\quad Depth & 79.0 [71.0, 85.3] & 80.6 [72.8, 86.6] & \textcolor{ForestGreen}{+1.6} & 0.8 \\
\quad Spatial & 88.8 [82.6, 93.0] & 88.8 [82.6, 93.0] & +0.0 & 0.8 \\
\midrule
\multicolumn{5}{l}{\textbf{Qwen2.5-VL-3B}} \\
\quad Forensic & 31.1 [23.8, 39.4] & 40.2 [32.2, 48.7] & \textcolor{ForestGreen}{+9.1} & 0.4 \\
\quad Vis.\ Sim. & 55.6 [47.1, 63.7] & 65.9 [57.6, 73.4] & \textcolor{ForestGreen}{+10.4} & 0.4 \\
\quad Art Style & 47.0 [38.2, 56.0] & 54.7 [45.7, 63.4] & \textcolor{ForestGreen}{+7.7} & 0.1 \\
\quad Counting & 68.3 [59.6, 76.0] & 72.5 [63.9, 79.7] & \textcolor{ForestGreen}{+4.2} & 0.3 \\
\quad Depth & 53.2 [44.5, 61.8] & 70.2 [61.6, 77.5] & \textcolor{ForestGreen}{+16.9} & 0.2 \\
\quad Spatial & 77.6 [70.1, 83.7] & 81.1 [73.9, 86.7] & \textcolor{ForestGreen}{+3.5} & 0.3 \\
\midrule
\multicolumn{5}{l}{\textbf{InternVL2.5-8B}} \\
\quad Forensic & 27.3 [20.4, 35.4] & 30.3 [23.1, 38.6] & \textcolor{ForestGreen}{+3.0} & 0.2 \\
\quad Vis.\ Sim. & 83.7 [76.6, 89.0] & 85.2 [78.2, 90.2] & \textcolor{ForestGreen}{+1.5} & 0.2 \\
\quad Art Style & 72.7 [63.9, 79.9] & 76.1 [67.6, 82.9] & \textcolor{ForestGreen}{+3.4} & 0.4 \\
\quad Counting & 72.5 [63.9, 79.7] & 72.5 [63.9, 79.7] & +0.0 & 0.6 \\
\quad Depth & 69.3 [60.8, 76.8] & 78.2 [70.2, 84.6] & \textcolor{ForestGreen}{+8.9} & 0.6 \\
\quad Spatial & 81.8 [74.7, 87.3] & 83.9 [77.0, 89.0] & \textcolor{ForestGreen}{+2.1} & 0.8 \\
\midrule
\multicolumn{5}{l}{\textbf{LLaVA-OV-7B}} \\
\quad Forensic & 35.6 [28.0, 44.1] & 45.5 [37.2, 54.0] & \textcolor{ForestGreen}{+9.8} & 0.6 \\
\quad Vis.\ Sim. & 70.4 [62.2, 77.4] & 81.5 [74.1, 87.1] & \textcolor{ForestGreen}{+11.1} & 0.4 \\
\quad Art Style & 71.8 [63.0, 79.2] & 71.8 [63.0, 79.2] & +0.0 & 0.6 \\
\quad Counting & 70.8 [62.2, 78.2] & 72.5 [63.9, 79.7] & \textcolor{ForestGreen}{+1.7} & 0.8 \\
\quad Depth & 74.2 [65.8, 81.1] & 78.2 [70.2, 84.6] & \textcolor{ForestGreen}{+4.0} & 0.6 \\
\quad Spatial & 82.5 [75.5, 87.9] & 83.9 [77.0, 89.0] & \textcolor{ForestGreen}{+1.4} & 0.6 \\
\midrule
\multicolumn{5}{l}{\textbf{Idefics3-8B}} \\
\quad Forensic & 30.3 [23.1, 38.6] & 31.1 [23.8, 39.4] & \textcolor{ForestGreen}{+0.8} & 0.0 \\
\quad Vis.\ Sim. & 57.8 [49.3, 65.8] & 68.1 [59.9, 75.4] & \textcolor{ForestGreen}{+10.4} & 0.0 \\
\quad Art Style & 62.4 [53.4, 70.6] & 65.8 [56.8, 73.8] & \textcolor{ForestGreen}{+3.4} & 0.8 \\
\quad Counting & 57.5 [48.6, 66.0] & 62.5 [53.6, 70.6] & \textcolor{ForestGreen}{+5.0} & 0.8 \\
\quad Depth & 61.3 [52.5, 69.4] & 64.5 [55.8, 72.4] & \textcolor{ForestGreen}{+3.2} & 0.5 \\
\quad Spatial & 85.3 [78.6, 90.2] & 85.3 [78.6, 90.2] & +0.0 & 0.5 \\
\bottomrule
\end{tabular}
\end{table}

\paragraph{Cross-model transfer of interpolation strength.} The sweeps above select settings separately for each model; the shared fixed protocol uses $\alpha_\text{interp}=0.4$, while calibration is evaluated separately. To test cross-model transfer, we ran leave-one-model-out selection over the seven sweeps: for each held-out model, we selected the $\alpha_\text{interp}$ that maximized mean CD delta on the other six and then applied it to the held-out model. The mean held-out delta is $-1.4\%$; the two-sided Wilcoxon test gives $p=0.22$ (the one-sided test for positive recovery gives $p=0.92$), and no single global $\alpha_\text{interp}$ is positive on average at any grid point. These results motivate prospective per-model calibration.

\clearpage

\section{Ablation Studies}
\label{app:ablations}

\subsection{Layer Sweep}
\label{app:layers}

\begin{figure}[!htbp]
    \centering
    \includegraphics[width=0.9\columnwidth]{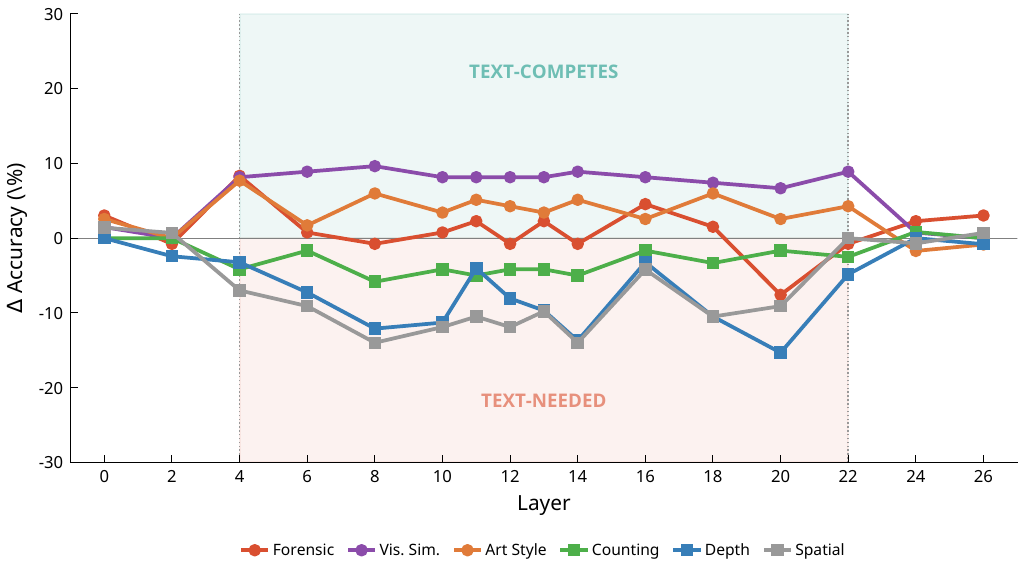}
    \caption{\textbf{Per-task layer sweep (Qwen2.5-VL-7B).} Individual task CD deltas across 16 injection layers at $\alpha_\text{interp}$=0 and $\alpha_\text{cd}$=0.65, using the same L12 centroid bank and complementing the group means in Figure~\ref{fig:layer_sweep}. The shaded L4--L22 interval marks the range where the group means separate; individual task trajectories vary and do not all preserve the group-level direction at every layer.}
    \label{fig:supp_layer_sweep}
\end{figure}

\begin{figure}[!htbp]
    \centering
    \includegraphics[width=0.6\columnwidth]{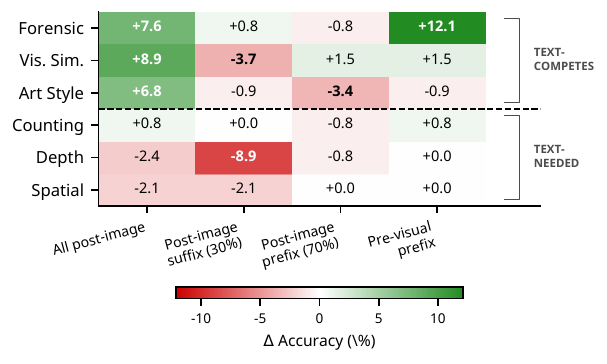}
    \caption{\textbf{Dose-dependent sensitivity across positional text spans.} CD effect when selectively applying centroid replacement to the full post-image tail, its first 70\% or last 30\%, or the pre-visual prefix (Qwen2.5-VL-7B, $\alpha_\text{interp}=0.4$, $\alpha_\text{cd}=1.0$). These ranges are positional heuristics rather than semantic parses. Full-post-image replacement produces the strongest gains on \textsc{text-competes} tasks (Forensic $+7.6\%$, Vis.\ Sim. $+8.9\%$, Art Style $+6.8\%$). Pre-visual-prefix replacement yields a large gain on Forensic Detection ($+12.1\%$). Suffix replacement at this dose is largely flat or mildly negative; at $\alpha_\text{interp}=0.3$ it recovers \textsc{text-competes} gains (\Cref{tab:segdose_tc}).}
    \label{fig:segment_ablation}
\end{figure}

We apply text centroid CD ($\alpha_\text{interp}$=0, $\alpha_\text{cd}$=0.65) at 16 layers (L0--L26) of Qwen2.5-VL-7B to test whether the separation is localized or distributed across depth, using the same L12 centroid bank rather than refitting at each layer. Figure~\ref{fig:supp_layer_sweep} shows per-task deltas. The \textsc{text-competes} group benefits from L4 through L22, while \textsc{text-needed} is hurt over much of the same range; effects shrink near the extremes. The separation is robust to injection layer across L4--L22. Because the groups were defined post hoc and the same L12 bank is used throughout, this sweep does not localize a mechanism.

\clearpage

\subsection{Specificity Controls}
\label{app:controls}

At the fixed primary protocol (Qwen2.5-VL-7B, L12, full post-image tail, $\alpha_\text{interp}=0.4$, $\alpha_\text{cd}=1$), we compare TCCD's nearest-centroid residual with an equal-norm random-direction reference. For each token, the random direction is scaled to exactly the residual norm, and both arms receive the same $(1-\alpha_\text{interp})$ scaling. With seed 42 set once for the deterministic run order, TCCD yields $+7.77\%$ across the three \textsc{text-competes} tasks versus $+0.24\%$ for matched random directions; across all six tasks, the means are $+3.27\%$ and $+0.40\%$ (\Cref{tab:matched_random}). This equal-norm control shows that perturbation direction matters beyond displacement magnitude alone. With one seed and one null distribution, however, it does not distinguish the specific role of $K$-means geometry.

\begin{table}[!htbp]
\centering
\caption{\textbf{Equal-norm matched-random specificity control (Qwen2.5-VL-7B).} Mean accuracy change (\%) at the fixed primary protocol: full post-image tail, L12, $\alpha_\text{interp}$=0.4, $\alpha_\text{cd}$=1.0, seed 42.}
\vspace{1em}
\label{tab:matched_random}
\small
\begin{tabular}{l r r}
\toprule
\textbf{Reference direction} & \textbf{\textsc{text-competes}} & \textbf{All six tasks} \\
\midrule
Nearest-centroid residual (TCCD) & \textbf{\textcolor{ForestGreen}{+7.77}} & \textbf{\textcolor{ForestGreen}{+3.27}} \\
Equal-norm random direction & +0.24 & +0.40 \\
\bottomrule
\end{tabular}
\end{table}

\subsection{The Full Positional-Span-by-Dose Grid}
\label{app:segdose}

\Cref{tab:segdose_tc} reports the complete positional-span $\times$ dose grid at $\alpha_\text{interp} \in \{0.2, 0.3, 0.4, 0.6\}$ under the primary protocol ($\alpha_\text{cd}$=1.0, L12, $K$=256, $n$=771). The implementation labels \texttt{question}, \texttt{options}, and \texttt{system} denote positional spans: the first 70\% of the post-image tail, its last 30\%, and the pre-visual prefix, respectively. The pre-visual range includes template and control tokens. All sources use the same pipeline and identical baselines; at $\alpha_\text{interp}$=0.4, post-image- and pre-visual-prefix cells reproduce the primary run to the fourth decimal. The effects vary with dose: pre-visual-prefix replacement is the only single range whose \textsc{text-competes} mean is positive at every dose (peaking at $+4.3\%$ at 0.4, with Forensic Detection from $+6.8\%$ to $+12.1\%$), suffix replacement contributes only at reduced doses ($+4.9\%$ to $+5.5\%$ at 0.2--0.3), and post-image-prefix replacement is null to mildly negative at every dose. No single range at any dose matches full-post-image replacement at its own optimum ($+7.8\%$ at 0.4). These results locate sensitivity by token position, not semantic role.

\begin{table}[!htbp]
\centering
\caption{\textbf{Positional-span-by-dose grid, summary (Qwen2.5-VL-7B).} Mean CD $\Delta$ (\%) by dose for \textsc{text-competes} (top) and all six tasks (bottom). Values come from the primary full-tail sweep, suffix dose-response control, and prefix positional-grid evaluation, all using the same pipeline.}
\vspace{1em}
\label{tab:segdose_tc}
\small
\begin{tabular}{l r r r r}
\toprule
& $\alpha_\text{interp}$=0.2 & $\alpha_\text{interp}$=0.3 & $\alpha_\text{interp}$=0.4 & $\alpha_\text{interp}$=0.6 \\
\midrule
\multicolumn{5}{l}{\textsc{text-competes} mean (3 tasks)} \\
    \quad Full post-image tail & +3.8 & +0.5 & \textbf{+7.8} & $-$4.2 \\
    \quad Post-image suffix (30\%) & +4.9 & \textbf{+5.5} & $-$1.3 & $-$4.3 \\
    \quad Post-image prefix (70\%) & $-$1.7 & $-$1.5 & $-$0.9 & $-$0.6 \\
    \quad Pre-visual prefix & +1.4 & +2.1 & \textbf{+4.3} & +2.5 \\
\midrule
\multicolumn{5}{l}{All-task mean (6 tasks)} \\
    \quad Full post-image tail & $-$4.4 & $-$5.0 & +3.3 & $-$1.9 \\
    \quad Post-image suffix (30\%) & $-$0.6 & $-$0.2 & $-$2.5 & $-$2.6 \\
    \quad Post-image prefix (70\%) & $-$1.1 & $-$1.3 & $-$0.7 & $-$0.3 \\
    \quad Pre-visual prefix & $-$0.1 & +1.1 & +2.3 & +1.4 \\
\bottomrule
\end{tabular}
\end{table}

\begin{table}[!htbp]
\centering
\caption{\textbf{Positional-span-by-dose grid, per-task detail} (post-image-prefix and pre-visual-prefix replacement; $\Delta$ in \%). $^*$ marks nominal, unadjusted McNemar $p<0.05$ with no multiplicity correction. Post-image-suffix summaries are in \Cref{tab:segdose_tc}; full-post-image cells are in the $\alpha_\text{interp}$ sweep of \Cref{app:alpha}.}
\vspace{1em}
\label{tab:segdose_detail}
\small
\begin{tabular}{l rrrr}
\toprule
\textbf{Task} & $\alpha_\text{interp}$=0.2 & $\alpha_\text{interp}$=0.3 & $\alpha_\text{interp}$=0.4 & $\alpha_\text{interp}$=0.6 \\
\midrule
\multicolumn{5}{l}{\textbf{Post-image prefix (first 70\%)}} \\
\quad Forensic & \textcolor{black}{+0.0} & \textcolor{red}{$-$0.8} & \textcolor{red}{$-$0.8} & \textcolor{ForestGreen}{+1.5} \\
\quad Vis.\ Sim. & \textcolor{black}{+0.0} & \textcolor{ForestGreen}{+1.5} & \textcolor{ForestGreen}{+1.5} & \textcolor{black}{+0.0} \\
\quad Art Style & \textcolor{red}{$-$5.1}$^*$ & \textcolor{red}{$-$5.1}$^*$ & \textcolor{red}{$-$3.4}$^*$ & \textcolor{red}{$-$3.4} \\
\quad Counting & \textcolor{red}{$-$0.8} & \textcolor{red}{$-$1.7} & \textcolor{red}{$-$0.8} & \textcolor{ForestGreen}{+0.8} \\
\quad Depth & \textcolor{black}{+0.0} & \textcolor{black}{+0.0} & \textcolor{red}{$-$0.8} & \textcolor{ForestGreen}{+0.8} \\
\quad Spatial & \textcolor{red}{$-$0.7} & \textcolor{red}{$-$1.4} & \textcolor{black}{+0.0} & \textcolor{red}{$-$1.4} \\
\midrule
\multicolumn{5}{l}{\textbf{Pre-visual prefix}} \\
\quad Forensic & \textcolor{ForestGreen}{+7.6} & \textcolor{ForestGreen}{+9.1}$^*$ & \textcolor{ForestGreen}{+12.1}$^*$ & \textcolor{ForestGreen}{+6.8}$^*$ \\
\quad Vis.\ Sim. & \textcolor{black}{+0.0} & \textcolor{ForestGreen}{+0.7} & \textcolor{ForestGreen}{+1.5} & \textcolor{ForestGreen}{+0.7} \\
\quad Art Style & \textcolor{red}{$-$3.4}$^*$ & \textcolor{red}{$-$3.4}$^*$ & \textcolor{red}{$-$0.9} & \textcolor{black}{+0.0} \\
\quad Counting & \textcolor{red}{$-$0.8} & \textcolor{black}{+0.0} & \textcolor{ForestGreen}{+0.8} & \textcolor{ForestGreen}{+0.8} \\
\quad Depth & \textcolor{red}{$-$2.4} & \textcolor{ForestGreen}{+0.8} & \textcolor{black}{+0.0} & \textcolor{red}{$-$0.8} \\
\quad Spatial & \textcolor{red}{$-$1.4} & \textcolor{red}{$-$0.7} & \textcolor{black}{+0.0} & \textcolor{ForestGreen}{+0.7} \\
\bottomrule
\end{tabular}
\end{table}

\subsection{\texorpdfstring{$N\times K$}{N x K} Sensitivity}
\label{app:ksweep}

We ran a held-out 30-cell $N\times K$ scaling grid (Qwen2.5-VL-7B, $\alpha_\text{cd}$=1.0, per-task best $\alpha_\text{interp}$). The surface is essentially flat: mean best $\Delta=5.2\%\pm0.4\%$ across all 30 cells with $N \in \{1{,}000, 2{,}000, 5{,}000, 10{,}000, 20{,}000, 50{,}000\}$ and $K \in \{128, 256, 512, 1{,}024, 2{,}048\}$, with no monotone trend in either direction. The signal is robust to both hyperparameters in this oracle-selected grid; we use $N$=2{,}000 and $K$=256 for cost.

\clearpage

\subsection{Sensitivity to Sink and Dead Tokens}
\label{app:sinks}

High-norm ``sink'' tokens~\citep{kang2025see, luo2025sink} and near-zero-norm ``dead'' tokens~\citep{kang2025see} could add noise to centroid fitting and inflate the apparent sufficiency of text centroids. We test this possibility on the primary checkpoint by refitting both visual and text centroids on Qwen2.5-VL-7B under three exclusion variants applied to the 2{,}000-image COCO cache: (i) \textsc{no\_dead} drops bottom-5\% norm tokens; (ii) \textsc{no\_sink} drops top-1\% norm tokens; (iii) \textsc{no\_either} applies both filters. \Cref{tab:sink_dead} reports the resulting mean visual-centroid cost, mean text-centroid cost, and mean oracle best-$\alpha_\text{interp}$ CD delta on BLINK 6 tasks.

\begin{table}[!htbp]
\centering
\caption{\textbf{Sensitivity to sink/dead-token exclusions (Qwen2.5-VL-7B, BLINK 6 tasks).} Centroids are refit on 2{,}000 COCO activations under three exclusion variants. \textsc{no\_dead} drops bottom-5\% norm tokens; \textsc{no\_sink} drops top-1\% norm tokens; \textsc{no\_either} applies both filters. Mean oracle best-$\alpha_\text{interp}$ CD delta changes by at most 0.68\% from baseline, and the text centroid cost is identical across all four variants.}
\vspace{1em}
\label{tab:sink_dead}
\small
\resizebox{\textwidth}{!}{%
\begin{tabular}{l r r r r r}
\toprule
\textbf{Variant} & \textbf{Vis.\ Kept} & \textbf{Text Kept} & \textbf{Vis.\ Cost} & \textbf{Text Cost} & \textbf{Mean Best $\Delta$} \\
\midrule
Baseline & 707{,}414 & 32{,}000 & $+1.4$ & $+27.2$ & \textcolor{ForestGreen}{$+5.4$} \\
\textsc{no\_dead} (bottom-5\% norm) & 672{,}043 & 30{,}400 & $+0.3$ & $+27.2$ & \textcolor{ForestGreen}{$+4.8$} \\
\textsc{no\_sink} (top-1\% norm) & 700{,}339 & 31{,}680 & $+1.7$ & $+27.2$ & \textcolor{ForestGreen}{$+5.0$} \\
\textsc{no\_either} (both) & 664{,}968 & 30{,}080 & $+2.3$ & $+27.2$ & \textcolor{ForestGreen}{$+5.2$} \\
\midrule
$\Delta$ vs baseline (\textsc{no\_dead}) & --- & --- & $-1.08$ & +0 & \textcolor{red}{$-0.68$} \\
$\Delta$ vs baseline (\textsc{no\_sink}) & --- & --- & $+0.26$ & +0 & \textcolor{red}{$-0.40$} \\
$\Delta$ vs baseline (\textsc{no\_either}) & --- & --- & $+0.89$ & +0 & \textcolor{red}{$-0.25$} \\
\bottomrule
\end{tabular}}
\end{table}

Across the four variants, mean oracle CD delta shifts by at most 0.68\%, while text cost remains +27.2\%. Visual cost moves from +1.4\% to +2.3\% under \textsc{no\_either}. Thus, the Qwen result is insensitive to the tested norm-tail exclusions. This control does not test whether sink tokens explain the negative visual costs observed for InternVL2.5-8B or Idefics3-8B.

\clearpage

\section{Statistical Analysis}
\label{app:stats}

\subsection{Confidence Intervals}
\label{app:ci}

\Cref{tab:supp_wilson} reports Wilson 95\% confidence intervals for all six tasks on Qwen2.5-VL-7B at the fixed protocol ($\alpha_\text{interp}$=0.4, $\alpha_\text{cd}$=1.0, $K$=256, $N$=2{,}000 COCO images). Greedy decoding produces deterministic predictions, so intervals reflect finite-sample binomial uncertainty rather than run-to-run variance. Visual Similarity (+8.9\%, $h$=0.227) exceeds Cohen's conventional ``small'' threshold of 0.2, and Forensic Detection (+7.6\%, $h$=0.152) and Art Style (+6.8\%, $h$=0.139) sit close to it. The remaining tasks show small or null effects at this fixed $\alpha_\text{interp}$; oracle per-task selection recovers additional gains (\Cref{tab:cd_results}).

\begin{table}[!htbp]
\centering
\caption{\textbf{Wilson 95\% confidence intervals (Qwen2.5-VL-7B, fixed $\alpha_\text{interp}$=0.4, $\alpha_\text{cd}$=1.0, $K$=256).} Baseline and +CD accuracies and $\Delta$ are in \%. Greedy decoding produces deterministic predictions; intervals reflect finite-sample binomial uncertainty, not run-to-run variance. Cohen's $h$ clears the conventional ``small'' threshold of 0.2 on Visual Similarity and sits near it on Forensic Detection and Art Style.}
\vspace{1em}
\label{tab:supp_wilson}
\small
\begin{tabular}{l r c c r r}
\toprule
\textbf{Task} & $n$ & \textbf{Base [95\% CI]} & \textbf{+CD [95\% CI]} & $\Delta$ & $h$ \\
\midrule
Forensic & 132 & 47.7 [39.4, 56.2] & 55.3 [46.8, 63.5] & \textcolor{ForestGreen}{+7.6} & .152 \\
Vis.\ Sim. & 135 & 76.3 [68.5, 82.7] & 85.2 [78.2, 90.2] & \textcolor{ForestGreen}{+8.9} & .227 \\
Art Style & 117 & 55.6 [46.5, 64.2] & 62.4 [53.4, 70.6] & \textcolor{ForestGreen}{+6.8} & .139 \\
\midrule
Counting & 120 & 66.7 [57.8, 74.5] & 67.5 [58.7, 75.2] & \textcolor{ForestGreen}{+0.8} & .018 \\
Depth & 124 & 79.0 [71.0, 85.3] & 76.6 [68.4, 83.2] & \textcolor{red}{$-$2.4} & $-$.058 \\
Spatial & 143 & 88.8 [82.6, 93.0] & 86.7 [80.2, 91.3] & \textcolor{red}{$-$2.1} & $-$.064 \\
\bottomrule
\end{tabular}
\end{table}

\subsection{Paired-Inference Scope}
\label{app:power}

Cohen's $h$ in \Cref{tab:supp_wilson} is a descriptive difference between two proportions. Because baseline and intervention predictions are paired on the same items, independent-group power calculations do not apply; paired power depends on the discordant cells. Wilson intervals describe marginal accuracies, while McNemar tests use paired changes. Tests reported after per-task $\alpha_\text{interp}$ selection are explicitly nominal and unadjusted. At fixed $\alpha_\text{interp}=0.4$, the six-task aggregate gives $p=0.025$, but no individual Qwen2.5-VL-7B task survives Bonferroni correction. Across the other checkpoints, five effects meet the two-sided per-model Bonferroni threshold: gains on Qwen2.5-VL-3B (Visual Similarity, $+10.4\%$) and Qwen3-VL-8B (Art Style, $+7.7\%$), and losses on Qwen3-VL-8B (Spatial Relation, $-9.8\%$) and Qwen3-VL-4B (Forensic Detection, $-20.5\%$; Spatial Relation, $-12.6\%$). The fixed protocol therefore yields both gains and harms across models, consistent with \Cref{tab:mci}.

\subsection{Fixed vs.\ Oracle Interpolation Strength}
\label{app:fixed_alpha}

We report the fixed protocol ($\alpha_\text{interp}$=0.4) separately from the same-set per-task optimum, which is an oracle analysis. \Cref{tab:supp_fixed_vs_best} compares them on Qwen2.5-VL-7B. The fixed-protocol mean is +3.3\%, the oracle mean is +5.4\%, and their ratio is 0.60 using the unrounded means. Per-task selection increases the gain, while the fixed protocol preserves the same group-level pattern.

\clearpage

\begin{table}[!t]
\centering
\caption{\textbf{Fixed $\alpha_\text{interp}$=0.4 vs.\ per-task oracle (Qwen2.5-VL-7B; $\Delta$ in \%).} Mean gains are +3.3\% at fixed $\alpha_\text{interp}$=0.4 and +5.4\% at the same-set per-task optimum; their ratio is 0.60 using unrounded means. The task-level pattern (\textsc{text-competes} $\gg$ \textsc{text-needed}) appears under both protocols.}
\vspace{1em}
\label{tab:supp_fixed_vs_best}
\small
\begin{tabular}{l r r r}
\toprule
\textbf{Task} & \textbf{Fixed 0.4} & \textbf{Oracle} & \textbf{$\alpha_\text{interp}^{*}$} \\
\midrule
Forensic & \textcolor{ForestGreen}{+7.6} & \textcolor{ForestGreen}{+11.4} & 0.5 \\
Vis.\ Sim. & \textcolor{ForestGreen}{+8.9} & \textcolor{ForestGreen}{+10.4} & 0.1 \\
Art Style & \textcolor{ForestGreen}{+6.8} & \textcolor{ForestGreen}{+6.8} & 0.4 \\
\midrule
Counting & \textcolor{ForestGreen}{+0.8} & \textcolor{ForestGreen}{+2.5} & 0.5 \\
Depth & \textcolor{red}{$-$2.4} & \textcolor{ForestGreen}{+1.6} & 0.8 \\
Spatial & \textcolor{red}{$-$2.1} & +0 & 0.8 \\
\midrule
\textbf{Mean} & \textbf{\textcolor{ForestGreen}{+3.3}} & \textbf{\textcolor{ForestGreen}{+5.4}} & --- \\
\bottomrule
\end{tabular}
\end{table}

\subsection{Centroid Fitting Variance}
\label{app:variance}

We quantify centroid-fit sensitivity in two ways. First, we refit centroids five times with the primary implementation, varying only $K$-means seed on the fixed COCO subset (data seed 1337, $N$=2{,}000, $K$=256, $\alpha_\text{cd}$=1.0). \Cref{tab:supp_variance} evaluates each refit at the per-task $\alpha_\text{interp}^{*}$ selected on the primary fit, and \Cref{tab:supp_variance_fixed} uses fixed $\alpha_\text{interp}$=0.4. Second, a 3$\times$5 grid (three data subsets by five $K$-means seeds) was computed with the same generic harvest prompt but a separate $K$=512 \texttt{scikit-learn} harvest/evaluation implementation. \Cref{tab:supp_variance_decomp} summarizes within-seed and across-seed standard deviations; these are sensitivity summaries, not fitted factorial variance components. The second grid uses a different $K$ and fitting backend from the primary $K$=256 \texttt{faiss} implementation, and its absolute deltas differ by as much as $\sim$4.5\%. We therefore use each implementation to assess its own fit-to-fit variation rather than compare their absolute effects.

Three observations hold for Qwen2.5-VL-7B. (1) At the fixed protocol, maximum unrounded per-task refit variation is $\sigma=1.83\%$; the separate 3$\times$5 grid has maximum reported $\sigma_\text{total}=1.53\%$, and the group ordering holds in all five primary refits and all 15 separate-implementation fits. (2) Frozen oracle $\alpha_\text{interp}^{*}$ is more sensitive (up to $\sigma=2.5\%$ on Forensic), with mean gains across refits from +3.8\% to +5.4\%. (3) Full text-replacement cost is +27.2\% across the five primary refits, while visual cost ranges from +1.0\% to +2.8\%; the resulting ratio ranges from 9.7$\times$ to 26.7$\times$ without changing direction. This analysis does not refit the other six models and therefore does not establish cross-model variance.

\begin{table}[!htbp]
\centering
\caption{\textbf{Refit stability of the reported per-task deltas (Qwen2.5-VL-7B, primary implementation, data seed 1337).} CD delta (\%) across five independent $K$-means fits (seeds 42, 800, 1337, 2024, 8320), evaluated at the frozen per-task $\alpha_\text{interp}^{*}$ selected on the primary fit. The seed-42 column reproduces \Cref{tab:cd_results}; $\alpha_\text{interp}$ is not reselected for each fit. Forensic Detection has the largest sensitivity ($\sigma$=2.5\%).}
\vspace{1em}
\label{tab:supp_variance}
\small
\begin{tabular}{l r r r r r r}
\toprule
\textbf{Task} & \textbf{km=42} & \textbf{km=800} & \textbf{km=1337} & \textbf{km=2024} & \textbf{km=8320} & $\sigma$ \\
\midrule
Forensic & \textcolor{ForestGreen}{+11.4} & \textcolor{ForestGreen}{+9.1} & \textcolor{ForestGreen}{+3.8} & \textcolor{ForestGreen}{+8.3} & \textcolor{ForestGreen}{+9.1} & 2.5 \\
Vis.\ Sim. & \textcolor{ForestGreen}{+10.4} & \textcolor{ForestGreen}{+10.4} & \textcolor{ForestGreen}{+9.6} & \textcolor{ForestGreen}{+9.6} & \textcolor{ForestGreen}{+10.4} & 0.4 \\
Art Style & \textcolor{ForestGreen}{+6.8} & \textcolor{ForestGreen}{+6.8} & \textcolor{ForestGreen}{+6.8} & \textcolor{ForestGreen}{+4.3} & \textcolor{ForestGreen}{+7.7} & 1.2 \\
Counting & \textcolor{ForestGreen}{+2.5} & \textcolor{ForestGreen}{+2.5} & \textcolor{ForestGreen}{+3.3} & \textcolor{ForestGreen}{+2.5} & \textcolor{ForestGreen}{+3.3} & 0.4 \\
Depth & \textcolor{ForestGreen}{+1.6} & \textcolor{ForestGreen}{+0.8} & \textcolor{red}{$-$0.8} & \textcolor{red}{$-$1.6} & \textcolor{ForestGreen}{+0.8} & 1.2 \\
Spatial & +0 & +0 & +0 & +0 & +0 & 0.0 \\
\midrule
\textbf{Mean} & \textcolor{ForestGreen}{+5.4} & \textcolor{ForestGreen}{+4.9} & \textcolor{ForestGreen}{+3.8} & \textcolor{ForestGreen}{+3.9} & \textcolor{ForestGreen}{+5.2} & --- \\
\bottomrule
\end{tabular}
\end{table}

\begin{table}[H]
\centering
\caption{\textbf{Refit stability at the fixed deployment protocol (Qwen2.5-VL-7B, primary implementation, $\alpha_\text{interp}$=0.4).} Same five refits as \Cref{tab:supp_variance}; the seed-42 column reproduces \Cref{tab:supp_wilson}. The maximum unrounded per-task $\sigma$ is 1.83\%, and the \textsc{text-competes} $>$ \textsc{text-needed} group separation holds in all five refits.}
\vspace{1em}
\label{tab:supp_variance_fixed}
\small
\begin{tabular}{l r r r r r r}
\toprule
\textbf{Task} & \textbf{km=42} & \textbf{km=800} & \textbf{km=1337} & \textbf{km=2024} & \textbf{km=8320} & $\sigma$ \\
\midrule
Forensic & \textcolor{ForestGreen}{+7.6} & \textcolor{ForestGreen}{+11.4} & \textcolor{ForestGreen}{+12.1} & \textcolor{ForestGreen}{+11.4} & \textcolor{ForestGreen}{+8.3} & 1.8 \\
Vis.\ Sim. & \textcolor{ForestGreen}{+8.9} & \textcolor{ForestGreen}{+7.4} & \textcolor{ForestGreen}{+8.2} & \textcolor{ForestGreen}{+8.2} & \textcolor{ForestGreen}{+8.2} & 0.5 \\
Art Style & \textcolor{ForestGreen}{+6.8} & \textcolor{ForestGreen}{+6.8} & \textcolor{ForestGreen}{+6.8} & \textcolor{ForestGreen}{+4.3} & \textcolor{ForestGreen}{+7.7} & 1.2 \\
Counting & \textcolor{ForestGreen}{+0.8} & \textcolor{ForestGreen}{+2.5} & \textcolor{ForestGreen}{+2.5} & \textcolor{ForestGreen}{+1.7} & \textcolor{ForestGreen}{+2.5} & 0.7 \\
Depth & \textcolor{red}{$-$2.4} & \textcolor{red}{$-$3.2} & \textcolor{red}{$-$3.2} & \textcolor{red}{$-$2.4} & \textcolor{red}{$-$3.2} & 0.4 \\
Spatial & \textcolor{red}{$-$2.1} & \textcolor{red}{$-$0.7} & \textcolor{red}{$-$0.7} & \textcolor{red}{$-$1.4} & \textcolor{red}{$-$2.1} & 0.6 \\
\midrule
\textbf{Mean} & \textcolor{ForestGreen}{+3.3} & \textcolor{ForestGreen}{+4.0} & \textcolor{ForestGreen}{+4.3} & \textcolor{ForestGreen}{+3.6} & \textcolor{ForestGreen}{+3.6} & --- \\
\bottomrule
\end{tabular}
\end{table}

\begin{table}[H]
\centering
\caption{\textbf{Sensitivity summary over the 3$\times$5 fit grid} (fixed $\alpha_\text{interp}$=0.4, $\alpha_\text{cd}$=1.0). $\sigma_\text{kmeans}$ is the within-data-seed SD across five $K$-means seeds, averaged over three data seeds; $\sigma_\text{data}$ is the across-data-seed SD averaged over $K$-means seeds; $\sigma_\text{total}$ is the SD across all 15 fits. These descriptive SDs are not formal factorial variance components.}
\vspace{1em}
\label{tab:supp_variance_decomp}
\small
\begin{tabular}{l r r r}
\toprule
\textbf{Task} & $\sigma_\text{kmeans}$ & $\sigma_\text{data}$ & $\sigma_\text{total}$ \\
\midrule
Forensic & 1.4 & 1.3 & 1.5 \\
Vis.\ Sim. & 0.5 & 0.4 & 0.5 \\
Art Style & 0.9 & 0.8 & 1.0 \\
Counting & 0.6 & 0.4 & 0.7 \\
Depth & 0.5 & 0.6 & 0.8 \\
Spatial & 0.5 & 0.5 & 0.6 \\
\midrule
\textbf{Max} & 1.4 & 1.3 & \textbf{1.5} \\
\bottomrule
\end{tabular}
\end{table}

\clearpage

\section{Additional Analysis}
\label{app:additional}

\subsection{Calibration}
\label{app:calibration}

\Cref{tab:supp_calibration} reports expected calibration error (ECE) before and after text centroid CD on Qwen2.5-VL-7B with an earlier exploratory protocol and 11{,}218-image held-out COCO+ScienceQA cache ($K$=512, $\alpha_\text{interp}$=0.4, $\alpha_\text{cd}$=0.65; 10 equal-width bins). ECE increases on five of six tasks (mean $\Delta$ECE $=+0.017$) as mean confidence rises from 0.785 to 0.841. Under the primary protocol on the full six-task set ($n$=771, 15 equal-width bins), TCCD at $\alpha_\text{interp}$=0.4 and $\alpha_\text{cd}$=1 raises aggregate ECE from 0.105 to 0.133 ($\Delta$ECE $=+0.028$) while raising pooled accuracy by 3.2\%. Applying TCCD only to the least-confident 25\% of items by top-2 margin holds $\Delta$ECE to $+0.006$ while retaining $+3.0\%$ accuracy; at 50\% coverage, $\Delta$ECE is $+0.022$ with $+3.5\%$. The top-2 margin and these coverage levels were selected in the same sweep rather than on held-out data, and temperature scaling was not tested.

\begin{table}[!h]
\centering
\caption{\textbf{Calibration analysis under the earlier exploratory protocol} (Qwen2.5-VL-7B, 11{,}218-image held-out COCO+ScienceQA cache, $K$=512, $\alpha_\text{interp}$=0.4, $\alpha_\text{cd}$=0.65; 10 equal-width bins). Text centroid CD increases ECE on five of six tasks (mean $\Delta$ECE $=+0.017$).}
\vspace{1em}
\label{tab:supp_calibration}
\small
\begin{tabular}{l r r r r r}
\toprule
\textbf{Task} & \textbf{ECE\textsubscript{B}} & \textbf{ECE\textsubscript{CD}} & $\Delta$\textbf{ECE} & \textbf{Conf.\textsubscript{B}} & \textbf{Conf.\textsubscript{CD}} \\
\midrule
Forensic & .103 & .110 & \textcolor{red}{+.007} & .522 & .620 \\
Vis.\ Sim. & .093 & .085 & \textcolor{ForestGreen}{$-$.008} & .856 & .907 \\
Art Style & .277 & .299 & \textcolor{red}{+.022} & .833 & .880 \\
\midrule
Counting & .158 & .185 & \textcolor{red}{+.027} & .820 & .874 \\
Depth & .024 & .061 & \textcolor{red}{+.038} & .785 & .845 \\
Spatial & .037 & .055 & \textcolor{red}{+.018} & .894 & .919 \\
\midrule
\textbf{Mean} & .115 & .133 & +.017 & .785 & .841 \\
\bottomrule
\end{tabular}
\end{table}

\clearpage

\subsection{Cross-Task Centroid Transfer}
\label{app:transfer}

\Cref{tab:supp_transfer} compares four centroid sources under two different protocols: COCO (primary, $N$=2{,}000, $K$=256, $\alpha_\text{cd}$=1, per-task best $\alpha_\text{interp}$) and earlier exploratory $K$=512 fits from \textsc{competes}-only, \textsc{needed}-only, or all six BLINK tasks (fixed $\alpha_\text{interp}$=0.4, $\alpha_\text{cd}$=0.65). Task-derived banks use unlabeled input activations from the first 50 evaluation items in each contributing task and overlap the evaluation set, making the comparison transductive. COCO gives the largest mean (+5.4\% vs.\ +1.1\%/+1.3\%/+0.4\%), but the protocols also differ in $K$, fit size, evaluation pipeline, contrastive strength, interpolation selection, and fit--evaluation overlap. The table therefore does not isolate a general-geometry effect. COCO fitting uses no BLINK labels or evaluation prompts; possible image-level source overlap was not measured.

\begin{table}[H]
\centering
\caption{\textbf{Cross-task centroid-source comparison (Qwen2.5-VL-7B; $\Delta$ in \%).} COCO uses the primary protocol ($N$=2{,}000, $K$=256, $\alpha_\text{cd}$=1, per-task best $\alpha_\text{interp}$); task-derived columns use earlier exploratory $K$=512 fits from the first 50 evaluation inputs per task, at fixed $\alpha_\text{interp}$=0.4 and $\alpha_\text{cd}$=0.65. Because the protocols and fit/evaluation overlap differ, the comparison is descriptive.}
\vspace{1em}
\label{tab:supp_transfer}
\small
\begin{tabular}{l r r r r}
\toprule
\textbf{Task} & \textbf{COCO} & \textbf{COMP.} & \textbf{NEED.} & \textbf{All 6} \\
\midrule
Forensic & \textcolor{ForestGreen}{+11.4} & \textcolor{ForestGreen}{+2.3} & \textcolor{ForestGreen}{+7.6} & \textcolor{ForestGreen}{+2.3} \\
Vis.\ Sim. & \textcolor{ForestGreen}{+10.4} & \textcolor{ForestGreen}{+0.7} & \textcolor{ForestGreen}{+3.0} & \textcolor{ForestGreen}{+0.7} \\
Art Style & \textcolor{ForestGreen}{+6.8} & +0 & \textcolor{ForestGreen}{+0.9} & +0 \\
\midrule
Counting & \textcolor{ForestGreen}{+2.5} & \textcolor{ForestGreen}{+0.8} & \textcolor{red}{$-$0.8} & \textcolor{ForestGreen}{+1.7} \\
Depth & \textcolor{ForestGreen}{+1.6} & \textcolor{ForestGreen}{+4.0} & \textcolor{red}{$-$0.8} & +0 \\
Spatial & +0 & \textcolor{red}{$-$1.4} & \textcolor{red}{$-$2.1} & \textcolor{red}{$-$2.1} \\
\midrule
\textbf{Mean} & \textcolor{ForestGreen}{+5.4} & \textcolor{ForestGreen}{+1.1} & \textcolor{ForestGreen}{+1.3} & \textcolor{ForestGreen}{+0.4} \\
\bottomrule
\end{tabular}
\end{table}

\clearpage

\subsection{Cross-Benchmark Consistency}
\label{app:crossbench}

The text--visual asymmetry extends beyond BLINK, while recovery is more mixed: some oracle best-per-task settings improve accuracy, but positive recovery under non-oracle selection concentrates in BLINK. We evaluate five further benchmarks spanning visually-prompted perception (VPBench;~\citealp{vptblink2026fragile}), general perception (CV-Bench,~\citealp{tong2024cambrian}; MMStar,~\citealp{chen2024we}; MMVP,~\citealp{tong2024eyes}), and clinical perception (MedBLINK;~\citealp{asadi2026mirage}). \Cref{tab:cross_benchmark} summarizes them; \Cref{tab:vpbench_summary} and \Cref{tab:medblink_summary} provide per-model detail, including two clinically fine-tuned MedGemma models.

\paragraph{VPBench.} VPBench's visual markers can produce accuracy swings of up to 13\% under design changes~\citep{vptblink2026fragile}. Across 7 models $\times$ 2 tasks (depth: $n=1{,}730$; semantic correspondence: $n=1{,}023$, per model), aggregate text cost is +11.4\% and visual cost +3.7\% (3.1$\times$). Three per-task best-$\alpha_\text{interp}$ gains have nominal unadjusted McNemar $p<0.05$ after oracle selection. Because our hidden states are downstream of the visual markers, this experiment does not test robustness to marker design.

\paragraph{CV-Bench, MMStar, MMVP.} This exploratory sweep uses canonical answer scoring and measures text cost only. Per-model means over the four task cells (CV-Bench depth/distance, MMStar, MMVP) span +16\% to +29\%, while MMVP considered alone is near zero across all seven models. Mean CD gain is +0.3\% over 28 cells. Without visual costs, this sweep cannot establish a text--visual asymmetry.

\paragraph{MedBLINK.} MedBLINK covers 8 clinical perception tasks on seven general checkpoints, MedGemma-1.5-4B, and MedGemma-27B~\citep{sellergren2026medgemma15,sellergren2025medgemma}. Each model is evaluated on 1,283 items (task $n$: 200/141/134/200/200/88/176/144). Across 72 model--task points, aggregate text cost is +9.3\% and visual cost +4.6\% (2.0$\times$); the direction holds in 8 of 9 model averages, with LLaVA at 0.9$\times$. Thirteen positive best-per-task cells have nominal unadjusted McNemar $p<0.05$ after oracle selection; fixed $\alpha_\text{interp}=0.4$ is reported alongside. The largest auxiliary recovery is MedGemma-4B on image-enhancement detection: 50.00\%$\to$76.87\% (+26.87\%) at oracle $\alpha_\text{interp}=0.5$ ($n$=134; exact McNemar $b$=3, $c$=39, nominal $p=5.63\times10^{-9}$). As a post-selected result, it is an upper bound rather than evidence of fixed-dose or clinical-wide benefit. Both MedGemma checkpoints show text-dominant aggregate costs, although this cross-sectional comparison cannot attribute the pattern to clinical fine-tuning.

\paragraph{Additional forced-choice recovery checks.} These analyses use a space-prefixed answer-token scorer rather than the primary bare-token scorer, so we treat them as scoring sensitivities rather than primary-protocol accuracies. Qwen2.5-VL-7B gains +2.5\% under strict MMBench CircularEval on the 938 base questions with all four dataset rotations (76.5\%$\to$79.0\%; exact McNemar $b$=6, $c$=29, $p=1.2\times10^{-4}$), while the five-model MMBench mean is $-3.2\%$, with three nominally significant negative models. Fixed-setting results are null on MMMU ($-$0.2\%, $p$=0.93, $n$=847) and MathVista (+0.9\%, $p$=0.60, $n$=540).

\begin{table}[!htbp]
\centering
\caption{\textbf{Cross-benchmark summary (costs and deltas in \%).} Aggregate asymmetry is observed in every row where both modalities were measured. $^\dagger$The exploratory CV-Bench/MMStar/MMVP sweep measured text cost only and therefore does not establish an asymmetry. ``Wins'' are nominal $p<0.05$ counts after oracle best-$\alpha_\text{interp}$ selection and are not family-wise corrected.}
\vspace{1em}
\label{tab:cross_benchmark}
\small
\setlength{\tabcolsep}{4pt}
\begin{tabular}{l c r r c r r}
\toprule
\textbf{Benchmark} & \textbf{\#Models} & \textbf{Vis.\ Cost} & \textbf{Text Cost} & \textbf{Asym} & \textbf{\#Sig.\ Wins} & \textbf{\#Points} \\
\midrule
BLINK (primary) & 7/7 & $+6.5$ & $+25.9$ & 4$\times$ & 10 & 42 \\
CV-Bench+MMStar+MMVP$^\dagger$ & 7/7 & --- & $+23.2$ & --- & --- & 28 \\
VPBench & 7/7 & $+3.7$ & $+11.4$ & 3.1$\times$ & 3 & 14 \\
MedBLINK (general) & 7/7 & $+5.0$ & $+9.7$ & 1.9$\times$ & 11 & 56 \\
MedBLINK (medical) & 2/2 & $+3.1$ & $+8.0$ & 2.5$\times$ & 2 & 16 \\
\bottomrule
\end{tabular}
\end{table}

\begin{table}[H]
\centering
\caption{\textbf{VPBench summary (7 models $\times$ 2 tasks; baseline accuracies, costs, and deltas in \%).} Aggregate text cost is +11.4\% versus +3.7\% visual. Because marker variants were not tested, this table does not assess VPBench's known marker sensitivity. Best-$\alpha_\text{interp}$ deltas and nominal win counts use oracle selection.}
\vspace{1em}
\label{tab:vpbench_summary}
\small
\resizebox{\textwidth}{!}{%
\begin{tabular}{l c c r r c r r}
\toprule
\textbf{Model} & \textbf{Depth} & \textbf{Semantic} & \textbf{Vis.\ Cost} & \textbf{Text Cost} & \textbf{Asym} & \textbf{Best $\Delta$} & \textbf{\#Sig.} \\
\midrule
Qwen2.5-VL-7B (SFT) & 64.7 & 32.9 & $+0.6$ & $+11.0$ & 17.3$\times$ & \textcolor{red}{$-0.6$} & 0/2 \\
Qwen2.5-VL-3B (SFT) & 54.0 & 30.4 & $+4.5$ & $+4.7$ & 1.0$\times$ & \textcolor{ForestGreen}{$+3.5$} & 1/2 \\
Qwen3-VL-8B (SFT+) & 75.2 & 47.5 & $+10.5$ & $+23.5$ & 2.2$\times$ & +0 & 0/2 \\
Qwen3-VL-4B (SFT+) & 72.5 & 41.4 & $+7.9$ & $+19.2$ & 2.4$\times$ & \textcolor{ForestGreen}{$+2.3$} & 1/2 \\
InternVL2.5-8B (MPO) & 59.8 & 30.8 & +0 & $+7.3$ & vis $\approx$ 0 & \textcolor{ForestGreen}{$+1.0$} & 1/2 \\
LLaVA-OV-7B (DPO) & 67.6 & 30.6 & $+2.0$ & $+11.3$ & 5.6$\times$ & \textcolor{ForestGreen}{$+0.7$} & 0/2 \\
Idefics3-8B (SFT) & 55.4 & 25.7 & $+0.2$ & $+2.8$ & 12.1$\times$ & \textcolor{ForestGreen}{$+1.1$} & 0/2 \\
\bottomrule
\end{tabular}}
\end{table}

\begin{table}[H]
\centering
\caption{\textbf{MedBLINK summary (9 models $\times$ 8 tasks; baseline accuracy, costs, and deltas in \%).} Best-$\alpha_\text{interp}$ is an oracle upper bound; \#Sig. counts nominal, unadjusted $p<0.05$ wins after selection. Aggregate text cost exceeds visual cost for 8 of 9 models (LLaVA is the exception). $\dagger$\,The Idefics ratio uses $|\text{visual cost}|$. The MedGemma comparison is observational and does not isolate a fine-tuning effect.}
\vspace{1em}
\label{tab:medblink_summary}
\small
\resizebox{\textwidth}{!}{%
\begin{tabular}{l c r r c r r r}
\toprule
\textbf{Model} & \textbf{Mean Base} & \textbf{Vis.\ Cost} & \textbf{Text Cost} & \textbf{Asym} & \textbf{Best $\Delta$} & \textbf{Fixed $\Delta$} & \textbf{\#Sig.} \\
\midrule
Qwen2.5-VL-7B (SFT) & 54.3 & $+0.8$ & $+7.9$ & 10.4$\times$ & \textcolor{ForestGreen}{$+2.4$} & \textcolor{red}{$-3.7$} & 1/8 \\
Qwen2.5-VL-3B (SFT) & 52.0 & $+5.3$ & $+5.6$ & 1.1$\times$ & \textcolor{ForestGreen}{$+3.4$} & \textcolor{red}{$-0.1$} & 2/8 \\
Qwen3-VL-8B (SFT+) & 70.0 & $+11.0$ & $+23.8$ & 2.2$\times$ & \textcolor{ForestGreen}{$+3.2$} & \textcolor{red}{$-1.4$} & 1/8 \\
Qwen3-VL-4B (SFT+) & 65.6 & $+11.4$ & $+18.9$ & 1.7$\times$ & \textcolor{ForestGreen}{$+1.4$} & \textcolor{red}{$-2.2$} & 2/8 \\
InternVL2.5-8B (MPO) & 49.3 & $+0.6$ & $+2.9$ & 4.8$\times$ & \textcolor{ForestGreen}{$+3.2$} & \textcolor{red}{$-2.3$} & 1/8 \\
LLaVA-OV-7B (DPO) & 52.8 & $+7.0$ & $+6.5$ & 0.9$\times$ & \textcolor{ForestGreen}{$+3.2$} & \textcolor{red}{$-2.1$} & 2/8 \\
Idefics3-8B (SFT) & 48.5 & \textcolor{red}{$-0.7$} & $+1.9$ & 2.6$\times$\textsuperscript{$\dagger$} & \textcolor{ForestGreen}{$+3.8$} & \textcolor{ForestGreen}{$+0.9$} & 2/8 \\
\midrule
MedGemma-4B (clin.\ SFT) & 53.8 & $+0.7$ & $+7.5$ & 11.4$\times$ & \textcolor{ForestGreen}{$+5.1$} & \textcolor{red}{$-2.9$} & 1/8 \\
MedGemma-27B (clin.\ SFT, 8-bit) & 54.8 & $+5.6$ & $+8.5$ & 1.5$\times$ & \textcolor{ForestGreen}{$+3.8$} & \textcolor{red}{$-1.5$} & 1/8 \\
\bottomrule
\end{tabular}}
\end{table}

\clearpage

\subsection{Auxiliary Answer-Token Scoring Sensitivity Across 10 Models and 8 Benchmarks}
\label{app:grid_detail}

We also examine ten models for which the implementation has architecture-specific visual-token span finders (Qwen2.5-VL-7B/3B, Qwen3-VL-8B/4B, LLaVA-OV-7B, InternVL2.5-8B, Idefics3-8B, and Gemma-3-4B/12B/27B); the Qwen2-VL-7B fallback row is excluded. Seven MCQA benchmarks were scored with single space-prefixed letter tokens rather than the bare continuation tokens used in the primary evaluation. We therefore treat these as answer-scoring sensitivity results. Under this scorer, text cost exceeds visual cost in all 60 non-chance cells; MMVP's ten cells remain near chance. Raw logits and predictions were not retained, so primary-protocol rescoring requires a rerun. MME instead uses the protocol-matched binary yes/no scorer and gives nine strict text-cost orderings and one tie. Measurement used full replacement ($\alpha_\text{interp}=0$); recovery used fixed $\alpha_\text{interp}=0.4$ and $\alpha_\text{cd}=1$, with $n$=300--1{,}000 per cell.

\begin{table}[!htbp]
\centering
\caption{\textbf{Auxiliary answer-token-scoring sensitivity (10 models $\times$ 8 benchmarks).} Costs and recovery are reported in \%. Seven MCQA rows use the alternative scorer defined above and therefore measure scoring sensitivity rather than primary-protocol accuracy. MME uses protocol-matched binary scoring. Significance counts are nominal and unadjusted; discordant counts were not retained.}
\vspace{1em}
\label{tab:grid_summary}
\small
\begin{tabular}{l r c c r c}
\toprule
\textbf{Benchmark} & $n$ & \textbf{Text cost (\%)} & \textbf{Vis.\ cost (\%)} & \textbf{Recovery (\%)} & \textbf{Nominal sig.\ +/$-$} \\
\midrule
A-OKVQA & 1000 & +55 to +67 & 0 to +31 & $-$0.9 & 0 / 3 \\
SEED-Bench & 1000 & +38 to +52 & 0 to +36 & $-$0.6 & 0 / 2 \\
ScienceQA-IMG & 1000 & +37 to +62 & 0 to +9 & $-$0.3 & 0 / 2 \\
RealWorldQA & 436 & +17 to +39 & 0 to +19 & $-$0.9 & 0 / 0 \\
MMStar & 1000 & +19 to +38 & 0 to +27 & $-$1.1 & 0 / 3 \\
CV-Bench & 1000 & +5 to +45 & $-$1 to +22 & $-$5.1 & 1 / 8 \\
MME & 1000 & +23 to +39 & $-$1 to +31 & $-$2.7 & 1 / 6 \\
MMVP & 300 & $-$2 to +3 & $-$3 to +1 & +0.1 & 0 / 0 \\
\bottomrule
\end{tabular}
\end{table}

\Cref{tab:rwqa_detail} gives the full RealWorldQA result under the alternative scorer and omits ratios when visual cost is within one item of zero. On option-letter-free MME with the protocol-matched binary scorer, Qwen2.5-VL-7B has baseline balanced accuracy 0.857, text-replaced 0.500, and visual-replaced 0.858 ($n$=1{,}000; 500 yes/500 no). This pattern cannot be explained solely by A/B/C/D tokenization or a fixed yes bias. It does not, however, exclude intervention-induced response bias or disruption of task-critical text.

\begin{table}[!htbp]
\centering
\caption{\textbf{Auxiliary RealWorldQA sensitivity under the alternative answer-token scorer} across ten models with validated visual-token spans ($n$=436, full replacement, $\alpha_\text{interp}=0$; costs in \%). Ratios with visual cost within one item of zero are reported as ``vis $\approx$ 0.''}
\vspace{1em}
\label{tab:rwqa_detail}
\small
\begin{tabular}{l r r c}
\toprule
\textbf{Model} & \textbf{Text cost (\%)} & \textbf{Vis.\ cost (\%)} & \textbf{Ratio} \\
\midrule
Qwen2.5-VL-7B & $+36.2$ & $+1.8$ & 19.8$\times$ \\
Qwen3-VL-8B & $+38.3$ & $+15.8$ & 2.4$\times$ \\
Qwen2.5-VL-3B & $+30.5$ & $+19.3$ & 1.6$\times$ \\
Qwen3-VL-4B & $+39.0$ & $+14.7$ & 2.7$\times$ \\
LLaVA-OV-7B & $+31.9$ & $+0.5$ & 69.5$\times$ \\
InternVL2.5-8B & $+35.8$ & $+0.2$ & vis $\approx$ 0 \\
Idefics3-8B & $+24.5$ & $-0.2$ & vis $\approx$ 0 \\
Gemma-3-4B & $+16.5$ & $+0.7$ & 24$\times$ \\
Gemma-3-12B & $+26.6$ & $+16.7$ & 1.6$\times$ \\
Gemma-3-27B & $+27.1$ & $+18.1$ & 1.5$\times$ \\
\bottomrule
\end{tabular}
\end{table}

\subsection{Contrastive Decoding Baselines and Auxiliary Sensitivity}
\label{app:cd_baselines}

\Cref{tab:cd_baselines} compares methods using the primary answer scorer at fixed settings: TCCD +3.27\% ($\alpha_\text{interp}$=0.4), text-only LCD +2.43\%, VCD +2.03\%, DoLa-low/high +1.61\%/+3.42\%, and SDCD +0.87\%, all at $\alpha_\text{cd}$=1. These values come from separate implementations on the same six BLINK task rows. \Cref{tab:cd_baselines_detail} gives an additional sensitivity analysis. Its left block fixes $\alpha_\text{interp}$=0.4 and selects $\alpha_\text{cd}$ from $\{0.2,0.4,0.6,0.8,1.0\}$ under the alternative answer-token scorer; the auxiliary LCD arm uses a black-image proxy. Raw logits and predictions were not retained, so bare-continuation rescoring requires rerunning. The right block reports primary-scored LCD/VCD on the other six models at fixed $\alpha_\text{cd}$=1. Methods are comparable within each block, not across them.

Under the alternative scorer, retaining one dataset row per MMBench base ID gives TCCD/LCD/VCD/DoLa flip-indicator correlations of $r$=0.46/0.41/0.61 ($n$=1{,}176); TCCD and LCD are uniquely correct on 27 and 30 items, respectively. Their oracle union reaches 86.4\% versus 84.1\% for the better method (baseline 82.9\%), a +2.3\% label-dependent upper bound. In a separate primary-scored BLINK portfolio using blank-image LCD, TCCD and LCD are uniquely correct on 36 and 37 items, the oracle union reaches 77.6\% versus 72.9\% (+4.7\%), and an option-letter majority vote realizes +1.3\%. These oracle unions are label-dependent upper bounds rather than deployable selection rules.

On alternative-scored MMBench, the same vote gains 1.0\% over baseline but remains 0.2\% below the better single method. The restricted CircularEval sensitivity uses the 938 bases with all four canonical dataset rotations: baseline 718/938, TCCD 741/938, exact McNemar $b=6$, $c=29$, $p=1.17\times10^{-4}$. It excludes 53 three-rotation bases because a constructed fourth rotation would move label-referential fixed options; a custom four-way treatment of all 991 bases gives +2.8\%. These MMBench results use canonical row groupings but not the primary answer-token scorer.

An auxiliary OPERA-inspired~\citep{huang2024opera} per-letter attention-penalty screen uses the alternative scorer on five tasks ($n$=8 each). Mean $\Delta$ is zero for every $\lambda \in \{0.5,1,5,10,50\}$: all cells are unchanged through $\lambda=10$, and two opposite one-item flips cancel at $\lambda=50$. This small greedy single-token screen is not an implementation of OPERA, which applies its penalty during beam search; a generation-based comparison would be required.

\begin{table}[!htbp]
\centering
\caption{\textbf{Extended CD baselines.} Left: Qwen2.5-VL-7B under the alternative answer scorer at fixed $\alpha_\text{interp}$=0.4, selecting $\alpha_\text{cd}$ at 0.4, by leave-one-task-out CV, or per-task oracle from $\{0.2,\ldots,1.0\}$. Right: primary-scored LCD/VCD on the other six models at fixed $\alpha_\text{cd}$=1 (mean $\Delta$ in \%, six BLINK tasks, $n$=771 each). Compare methods only within each block.}
\vspace{1em}
\label{tab:cd_baselines_detail}
\small
\begin{tabular}{l r r r}
\toprule
\textbf{Qwen2.5-VL-7B (alt.)} & \textbf{$\alpha_\text{cd}$=.4} & \textbf{CV-$\alpha_\text{cd}$} & \textbf{Oracle} \\
\midrule
TCCD (ours) & $+1.65$ & $+1.67$ & $+3.22$ \\
Blank-image LCD & $+2.68$ & $+2.55$ & $+4.54$ \\
VCD & $-0.52$ & $-1.17$ & $+0.27$ \\
\bottomrule
\end{tabular}
\hspace{1.5em}
\begin{tabular}{l r r}
\toprule
\textbf{Model} & \textbf{LCD} & \textbf{VCD} \\
\midrule
Qwen2.5-VL-3B & $+5.18$ & $+5.43$ \\
Qwen3-VL-8B & $+3.35$ & $-0.97$ \\
Qwen3-VL-4B & $+1.04$ & $-2.97$ \\
InternVL2.5-8B & $-0.01$ & $+0.63$ \\
LLaVA-OV-7B & $+2.74$ & $-1.09$ \\
Idefics3-8B & $+0.45$ & $+1.77$ \\
\bottomrule
\end{tabular}
\end{table}

\subsection{Answer-Logit Resampling: Per-Task Detail}
\label{app:votek}

\Cref{tab:selfconsistency} gives the mode-level summary and \Cref{tab:votek_detail} per-task deltas. Seed 42 is used; all option-letter draws come from one frozen forward-pass softmax at $T$=0.7 with no top-$p$ truncation, and ties favor the first inserted draw (greedy in mixed mode). This adds negligible model compute beyond one forward pass. Because all draws reuse one softmax, it is an answer-logit resampling control rather than conventional self-consistency across generation paths.

\begin{table}[!htbp]
\centering
\caption{\textbf{TCCD vs.\ seed-pinned answer-logit resampling (Qwen2.5-VL-7B).} Mean $\Delta$ vs.\ greedy (\%), six BLINK tasks ($n$=771). All votes reuse one frozen softmax ($\approx$1$\times$ model compute); TCCD uses two forward passes.}
\vspace{1em}
\label{tab:selfconsistency}
\small
\begin{tabular}{l c c c}
\toprule
\textbf{Method} & \textbf{Vote@2} & \textbf{Vote@5} & \textbf{Vote@10} \\
\midrule
Mixed draws ($\approx$1$\times$) & 0.0 & $-$0.1 & $-$0.2 \\
Stochastic draws ($\approx$1$\times$) & $-$3.2 & $-$3.4 & $-$0.7 \\
Greedy repeats ($\approx$1$\times$) & 0.0 & 0.0 & 0.0 \\
\midrule
Text centroid CD, fixed $\alpha_\text{interp}$=0.4 (2$\times$) & \multicolumn{3}{c}{\textcolor{ForestGreen}{+3.3}} \\
Text centroid CD, oracle best-per-task $\alpha_\text{interp}$ (2$\times$) & \multicolumn{3}{c}{\textcolor{ForestGreen}{+5.4}} \\
\bottomrule
\end{tabular}
\end{table}

\begin{table}[!htbp]
\centering
\caption{\textbf{Per-task answer-logit-resampling deltas} (\% vs.\ greedy, Qwen2.5-VL-7B). Mixed = one greedy plus $k-1$ stochastic draws; stochastic = $k$ draws.}
\vspace{1em}
\label{tab:votek_detail}
\small
\begin{tabular}{l rrr rrr}
\toprule
& \multicolumn{3}{c}{\textbf{Mixed}} & \multicolumn{3}{c}{\textbf{Pure-stochastic}} \\
\textbf{Task} & @2 & @5 & @10 & @2 & @5 & @10 \\
\midrule
Forensic & 0.00 & $-$0.76 & +0.76 & $-$9.09 & $-$4.55 & 0.00 \\
Vis.\ Sim. & 0.00 & +0.74 & $-$1.48 & $-$0.74 & $-$2.96 & +0.74 \\
Art Style & 0.00 & +0.85 & +0.85 & 0.00 & $-$1.71 & +0.85 \\
Counting & 0.00 & 0.00 & 0.00 & $-$4.17 & $-$0.83 & 0.00 \\
Depth & 0.00 & $-$0.81 & 0.00 & $-$2.42 & $-$6.45 & $-$3.23 \\
Spatial & 0.00 & $-$0.70 & $-$1.40 & $-$2.80 & $-$4.20 & $-$2.80 \\
\bottomrule
\end{tabular}
\end{table}

\subsection{Three-Stage Sufficiency Probe and Visual Coarsening Curve}
\label{app:stage_probe}

This exploratory Qwen2.5-VL-7B probe uses full replacement, $K$=256, and $n$=771 BLINK items. Raw-encoder and post-projector centroids were fit on 300 COCO images; the L16 point reuses the primary 2{,}000-image bank. Costs are 22\% at the raw vision-encoder output, 21\% after the projector, and 2\% at L16. A layer-fit sweep gives L0/L4/L8/L12/L16/L20/L24 costs of 24/23/22/21/2/0/$-$0\%. The drop suggests visual states become centroid-summarizable between L12 and L16, but stage, fit sample, and prompt context are not fully matched, so this is not a controlled localization of modal competition.

\subsection{All Fourteen BLINK Tasks}
\label{app:blink14}

\Cref{tab:blink14} reports fixed-protocol TCCD on all fourteen BLINK tasks for Qwen2.5-VL-7B and Qwen3-VL-8B. Across the eight remaining tasks, the descriptive mean of sixteen model--task deltas is $-$0.9\%; we do not pool a McNemar test across models because the same items recur. Multi-view Reasoning is not assigned to either task group and reaches an oracle +7.5\% on Qwen2.5-VL-7B ($\alpha_\text{interp}$=0.8; nominal unadjusted $p$=0.012 after selection). Functional Correspondence's fixed +12.3\% sits on a low 17.7\% baseline under the fixed four-letter readout and may therefore reflect a floor effect.

\begin{table}[!htbp]
\centering
\caption{\textbf{All fourteen BLINK tasks at fixed $\alpha_\text{interp}$=0.4} under the fixed four-letter readout (baseline and $\Delta$ in \%; McNemar $p$). Top: the six analyzed tasks; bottom: the remaining eight. Several Qwen3-VL-8B baselines in the lower block are substantially higher than the corresponding Qwen2.5-VL-7B baselines.}
\vspace{1em}
\label{tab:blink14}
\footnotesize
\setlength{\tabcolsep}{4pt}
\begin{tabular}{l r rrr rrr}
\toprule
& & \multicolumn{3}{c}{\textbf{Qwen2.5-VL-7B}} & \multicolumn{3}{c}{\textbf{Qwen3-VL-8B}} \\
\textbf{Task} & $n$ & Base (\%) & $\Delta$ (\%) & $p$ & Base (\%) & $\Delta$ (\%) & $p$ \\
\midrule
Forensic Detection & 132 & 47.7 & +7.6 & .096 & 66.7 & $-$6.1 & .011 \\
Visual Similarity & 135 & 76.3 & +8.9 & .014 & 87.4 & +1.5 & .317 \\
Art Style & 117 & 55.6 & +6.8 & .088 & 45.3 & +7.7 & .003 \\
Counting & 120 & 66.7 & +0.8 & .655 & 64.2 & $-$0.8 & .739 \\
Relative Depth & 124 & 79.0 & $-$2.4 & .491 & 88.7 & $-$4.8 & .034 \\
Spatial Relation & 143 & 88.8 & $-$2.1 & .491 & 87.4 & $-$9.8 & .006 \\
\midrule
Multi-view Reasoning & 133 & 55.6 & +0.0 & 1.000 & 55.6 & +0.0 & 1.000 \\
Jigsaw & 150 & 55.3 & $-$2.0 & .706 & 56.0 & $-$4.0 & .366 \\
Object Localization & 122 & 45.1 & $-$1.6 & .739 & 63.9 & +0.0 & 1.000 \\
Visual Correspondence & 172 & 32.6 & $-$2.9 & .446 & 77.3 & $-$2.9 & .025 \\
Semantic Correspondence & 139 & 32.4 & $-$2.9 & .493 & 50.4 & +1.4 & .480 \\
Relative Reflectance & 134 & 32.1 & +0.7 & .796 & 44.8 & $-$2.2 & .549 \\
Functional Correspondence & 130 & 17.7 & +12.3 & .011 & 42.3 & $-$4.6 & .221 \\
IQ Test & 150 & 19.3 & $-$2.0 & .655 & 26.7 & $-$4.0 & .157 \\
\bottomrule
\end{tabular}
\end{table}

\subsection{Symmetric Visual-Side Contrastive Decoding (VIS-CCD)}
\label{app:visccd}

The symmetric intervention contrasts a \emph{visual}-centroid-erased reference under the same protocol. At oracle best-per-task $\alpha_\text{interp}$, the VIS-CCD/TCCD mean-delta ratio is 0.24 on InternVL2.5-8B (+0.75\% vs.\ +3.15\%), 0.76 on LLaVA-OV-7B (+3.54\% vs.\ +4.68\%), and 1.03 on Qwen2.5-VL-7B (+5.59\% vs.\ +5.45\%). These post-selected comparisons show that intervention-level asymmetry is weaker and more model-dependent than measurement-level asymmetry, without a consistent advantage for the text-side intervention.

\subsection{Sighted--Blind Judge Contrast on Selected Flips}
\label{app:gemini}

For 79 successful TCCD flips selected under the auxiliary answer-token scorer (37 BLINK and 42 MMBench), two external judges answered each original MCQA item with and without the image. The May 2026 API run used \texttt{gemini-3.1-pro-preview} and \texttt{gemini-3.1-flash-lite} at temperature 1.0. Both prompts required A/B/C/D-only answers, and the embedded question and options were identical. Pro agreed with the corrected answer on 93.7\% of sighted calls versus 48.1\% of blind calls (per-source gaps: BLINK +43.2\%; MMBench +47.6\%); Flash-Lite gave 81.0\% versus 44.3\%. Sighted raw inter-judge agreement was 65/79, with reported $\kappa=0.645$; the paired 4$\times$4 table needed to reconstruct $\kappa$ was not retained. Sighted and blind calls also used different thinking levels and output-token caps. Because sighted and blind calls differ in both image availability and inference settings, and ground truth defines the selected flip set, this comparison supports the plausibility of the corrections but does not isolate image dependence or estimate net recovery. A complementary blank-image comparison is inconclusive: +2.4\% with the image versus +0.6\% without (interaction +1.8\%; one-sided per-sample sign test, 92 positive versus 78 negative, $p=0.159$, $n$=771).

\subsection{Open-Ended Generation and DocVQA}
\label{app:generative}

The generative checks use Qwen2.5-VL-7B and the primary centroid bank. On the first 2{,}000 OK-VQA validation items, first-token-only TCCD (text L12, $\alpha_\text{interp}=0.4$, $\alpha_\text{cd}=1$, greedy continuation up to 8 tokens) lowers canonical VQA soft accuracy by 6.6\% (paired $t(1999)=-8.454$, $p=5.33\times10^{-17}$); exact match falls 7.2\% (exact McNemar $b$=224, $c$=80, $p=6.06\times10^{-17}$). On 2{,}000 streamed COCO validation images, a second \texttt{generate} pass with the L12 replacement hook active throughout ($\alpha_\text{interp}=0.4$, up to 48 tokens, prompt ``Describe this image in a few sentences.'') lowers object recall by 2.9\% (paired $t(1999)=-5.799$, $p=7.72\times10^{-9}$) with $\Delta$CHAIR$_i=-0.001$. The captioning arm labeled \texttt{tccd} applies centroid-replaced generation rather than contrastive decoding and therefore does not use $\alpha_\text{cd}$. On DocVQA ($n$=300), greedy prefill-only full replacement at text L12 or visual L16 collapses ANLS (baseline 0.945; text 0.00; visual 0.06). A gold-string containment diagnostic gives visual cost 0.90 versus text cost 0.71, reversing the perception-competition pattern where the image carries the answer text. Together, these checks show no benefit in the tested generative and OCR settings.

\clearpage

\subsection{Method Positioning}
\label{app:positioning}

\Cref{tab:method_positioning} compares centroid replacement with the prior work discussed in \Cref{sec:related}, organized by each method's intervention locus and use. Centroid replacement operates on modality-specific hidden states at intermediate layers and serves first as a measurement instrument; TCCD uses the text-replaced pass as its contrastive reference.

\begin{table}[!htbp]
\centering
\caption{\textbf{Method positioning.} Centroid replacement operates on intermediate hidden-state centroids, used first as a measurement instrument for modal dependence; prior methods operate at other loci or are descriptive.}
\label{tab:method_positioning}
\footnotesize
\begin{tabular}{p{4.6cm} p{2.4cm} p{5.0cm}}
\toprule
\textbf{Method} & \textbf{Operates on} & \textbf{What it does} \\
\midrule
Centroid replacement / TCCD (ours) & hidden states (mid-stack) & measures text-vs-visual dependence; contrasts a text-erased pass \\
LCD~\citep{manevich2024mitigating} & output logits & contrasts the text-only LLM distribution \\
VCD~\citep{leng2024mitigating} & input pixels & contrasts a noise-distorted image \\
\citet{fu2025hiddenplainsightvlms} & analysis & visual information encoded but unused \\
\citet{wu2025mitigating} & training time & instance-level modality mixing \\
\citet{deng2025words}; \citet{li2025mbq} & behavior; quantization & text-over-vision bias; balanced quantization \\
\citet{huh2024platonic}; \citet{qi2025beyond}; \citet{papadimitriou2025interpreting} & representation geometry & cross-modal geometry, descriptive \\
\bottomrule
\end{tabular}
\end{table}

\section{LLM Disclosure}
LLMs assisted with prose editing, code review and implementation, and plotting author-generated data. The authors determined the scientific content and verified the resulting prose, figures, references, software changes, and experimental results. LLMs were also used as external judges only in the post-hoc Gemini analysis in \Cref{app:gemini}; otherwise, they did not generate experimental data or determine the paper's claims.

\end{document}